\documentclass[preprint,a4paper]{elsarticle}

\usepackage[hidelinks]{hyperref}
\usepackage{lineno,amsmath,amsfonts}
\journal{Journal of Mechanical Systems and Signal Processing}

\usepackage[capitalise]{cleveref}
\Crefformat{equation}{#2Eq.\ (#1)#3}
\Crefrangeformat{equation}{#2Eqs.\ (#1)#3}
\crefformat{appendix}{{#2Appendix #1#3}}
\usepackage{graphicx,subcaption}
\graphicspath{{./Figures/Experiment/}{./Figures/kernelDesign/}{./Figures/oneDimModel/}{./Figures/twoDimModel/}{./Figures/kernelDesign/}}
\usepackage[toc,page]{appendix}

\usepackage[none]{hyphenat}

\begin{document}

\begin{frontmatter}

% \title{Designing machine learning tools for guided wave modelling and understanding}
%\title{An Enhanced Machine Learning Approach For Modelling Guided Waves}
\title{Structured Machine Learning Tools for Modelling Characteristics of Guided Waves}
% Generation of Lamb Wave Feature-Space Models Using Gaussian Processes %
% Designing machine learning tools for guided wave modelling and understanding %

\author[1]{Marcus Haywood-Alexander\corref{cor1}}
\cortext[cor1]{Corresponding Author}
\ead{mhaywood-alexander1@sheffield.ac.uk}

\author[1]{Nikolaos Dervilis}

\author[1]{Keith Worden}

\author[1]{Elizabeth J.\ Cross}

\author[2]{Robin S.\ Mills}

\author[1]{Timothy J.\ Rogers}

\address[1]{Dynamics Research Group, Department of Mechanical Engineering, The University of Sheffield, Mappin Building, Mappin Street, Sheffield, S1 3JD, United Kingdom}
\address[2]{Laboratory for Verification and Validation (LVV), Europa Avenue, Sheffield S9 1ZA, United Kingdom}

\date{May 2020}

\begin{abstract}
    The use of ultrasonic guided waves to probe the materials/structures for damage continues to increase in popularity for non-destructive evaluation (NDE) and structural health monitoring (SHM). The use of high-frequency waves such as these offers an advantage over low-frequency methods from their ability to detect damage on a smaller scale. However, in order to assess damage in a structure, and implement any NDE or SHM tool, knowledge of the behaviour of a guided wave throughout the material/structure is important (especially when designing sensor placement for SHM systems). Determining this behaviour is extremely difficult in complex materials, such as fibre-matrix composites, where unique phenomena such as continuous mode conversion takes place. This paper introduces a novel method for modelling the feature-space of guided waves in a composite material. This technique is based on a data-driven model, where prior physical knowledge can be used to create structured machine learning tools; where constraints are applied to provide said structure. The method shown makes use of Gaussian processes, a full Bayesian analysis tool, and in this paper it is shown how physical knowledge of the guided waves can be utilised in modelling using an ML tool. This paper shows that through careful consideration when applying machine learning techniques, more robust models can be generated which offer advantages such as extrapolation ability and physical interpretation.
\end{abstract}


\end{frontmatter}

% \linenumbers

\section{Introduction}
\label{sec:intro}

In engineering applications, the use of complex materials, such as composite or porous materials can offer benefits thanks to their high strength-to-weight ratio \cite{Dowling2012}, amongst other advantages. However, with the increased usage of such materials, also comes the ability of building larger structures; an example being the ever-increasing size of wind turbines due to the use of glass-fibre reinforced polymer for the blades \cite{Schubel2012}. With larger structures also comes an increasing need for robust structural health monitoring systems in order to prolong life and reduce costs associated with maintenance, down time and repair of large-scale structures \cite{Wymore2015}. There are numerous strategies to perform structural health monitoring, many of which involve the use of frequency-based behaviour of the system \cite{Farrar2012}. Increasingly, high-frequency methods are being used, as they have an ability to detect defects on a smaller scale thanks to the analogy with diffraction theory, where a change in waves only occurs when they pass through diffractors of small enough size relative to their wavelength. One such example of these is to use \textit{ultrasonic guided waves} (UGWs), where a guided wave is induced within structure which acts as a wave-guide, and analysis of the `wave-packet' as it arrives in certain locations can give indications of inhomogeneities in the material. %

There have been many advances in the modelling of composite materials, such as in terms of the fatigue damage \cite{Mao2002}; a comparative review of state-of-the-art modelling methodologies for damage in composites has been made by Orifici \textit{et al.} \cite{Orifici2008}, in which they discuss many issues such as length scales and implicit modelling. Time-space modelling of guided waves can be solved analytically for isotropic materials \cite{Viktorov1967} or using numerical methods for layered composites \cite{Adler1990}; however, the complexity of these calculations becomes substantial when modelling the interaction with damage or for a fibre-composite. One primary issue with composites is the phenomena of \emph{continuous mode conversion} (CMC) \cite{Mook2014,Willberg2012}. This can be conceptualised as secondary guide behaviour of the fibres within the material, which for quasi-isotropic materials, with increased randomness of lamina orientation, it may be possible to model with a set lay-up configuration \cite{Wang2003}. %

Along with the increase in available computing power, recent years have shown a continuous increase in adoption of machine learning (ML) techniques applied to engineering modelling problems in order to overcome barriers in physic-based modelling, though not always successfully \cite{Simeone2018}. Many of these methods are applied to a dataset to generate a model of certain behaviour; however, this can limit applications. It is only reasonable to assume that a model fits the \textit{specific} material/structure, and generalisation cannot be implemented, so changes in design or environment would require retraining of the model. %

In this paper, a series of structured machine learning tools are presented in which physical knowledge is embedded by differing means. It is important to note this does not generate a directly interpretable model as material properties are not introduced or extracted from the methodology. Instead, belief is embedded from prior knowledge of the physics which control guided wave features, creating structured tools that are constrained by this physical knowledge. %

Using this novel method, study is done to investigate a novel path of understanding and predicting some behaviour characteristics of guided waves by utilising enhanced machine learning tools that capture the uncertainty of modelling the attenuation of such waves (important when designing ultrasonic monitoring systems on large scale structures like wind turbine blades). This paper focusses on one guided wave feature, the maximum amplitude of the Hilbert envelope of a wave-packet and although this paper focusses on features like energy attenuation, there is nothing to stop the implementation of this new view of modelling on other guided waves features.

This paper will discuss, briefly, the pertinent physics of guided waves in \cref{sec:UGW_physics} and then introduce the experimental setup used to demonstrate the methodology in \cref{sec:experiment}. The main contribution of this paper is presented in \cref{sec:models}; in this section, increasingly sophisticated models are generated which show how physical knowledge of the guided waves can help a machine learning strategy. This discussion begins with a purely physical one-dimensional attention model which is then extended to the two-dimensional case for modelling features across the surface of a composite plate. These models are compared, demonstrated and discussed in context of experimental data in \cref{sec:results}. Finally, conclusions are made in \cref{sec:conclusion} and possible future works are presented.

\section{Physics of guided waves}
\label{sec:UGW_physics}

Guided waves are used in several engineering applications, such as non-destructive evaluation (NDE) and structural health monitoring (SHM); with prior knowledge of how these waves behave, defects and inhomogeneities can be detected in structures. Such waves undergo an interesting phenomena when they occur in particular structures, such as rods, hollow cylinders and plates; they propagate primarily in the longitudinal direction perpendicular to oscillation and are known as guided waves. When such waves are guided due to propagation along the surface of a medium, they are called Rayleigh waves. However, if a wave travels in a bounded medium, perpendicular to two surfaces, where the wavelength is sufficiently long compared to the distance between these surfaces, often exhibited in plates, it is called a Lamb wave. Overviews of the derivations and characteristics of such waves are well described by Rose \cite{Rose2004} and Worden \cite{Worden2001}.

\subsection{Attenuation of guided waves}

As Rayleigh waves propagate along a surface of a structure, their amplitude $A$, decays with propagation distance \cite{Viktorov1967}, $x$, by 
\begin{equation}
    A(x) \propto \frac{1}{\sqrt{k_Rx}}
    \label{eq:Rayleigh_atten}
\end{equation}
where $k_R$ is the real wavenumber. The attenuation of Lamb waves depends on many factors, although Pollock \cite{Pollock1986} states the four most important ones to be:
\renewcommand{\labelenumi}{(\roman{enumi})}
\begin{enumerate}
    \item geometric spreading,
    \item material damping,
    \item dissipation into adjacent media,
    \item wave dispersion,
\end{enumerate}

Attenuation of Lamb waves has been accurately modelled by the inclusion of proportional damping through numerical and experimental studies \cite{Ramadas2011,Schubert2011}, and the effect of propagation distance on the amplitude of the wave has been described for geometric spreading \cite{Schubert2011} briefly,
\begin{equation}
    A(x) \propto A_0 \sqrt{x_0/x}
    \label{eq:Lamb_atten_geo}
\end{equation}
and material damping \cite{Rose2004} as,
\begin{equation}
    A(x) \propto A_0\exp\left(-\zeta_ix\right)
    \label{eq:Lamb_atten_damp}
\end{equation}
where $A_0$ \& $x_0$ are the amplitude and distance at an initial location from a point source, and $\zeta_i$ is the attenuation coefficient of the viscoelastic medium. Due to dispersive characteristics of Lamb waves, $\zeta_i$ is dependent on the central frequency of the wave. 
A key factor of guided waves in plates is the variety of wave modes that propagate within a single \textit{wave-packet}, which are split into two main types: symmetric, $S_n$, and antisymmetric, $A_n$, each of which has an increasing number of modes with increasing \textit{frequency-thickness} of the plates in question. 

\subsection{Anisotropic media and guided wave feature-spaces}

When Lamb waves propagate in anisotropic media, the modelling and solutions become very complex, even more so when attempting to model their interaction with defects \cite{Lowe1995, Alleyne1992, Guo1993}. A crucial characteristic of guided waves in fibrous materials such as carbon-fibre-reinforced-polymer (CFRP) is the phenomena of Continuous Mode Conversion (CMC) \cite{Pollock1986}, as shown by Mook \textit{et al} \cite{Mook2014} and Willberg \textit{et al} \cite{Willberg2012}, where the boundaries of layers or weaves cause conversion of $S_0$ modes into $A_0$ with frequent enough occurrence that they can be considered continuous along the propagation path. At propagation paths through the fibres, the energy of all modes is reduced thanks to this phenomena.

In order to use guided waves in damage detection strategies, a baseline state must first be determined, which is often done through the use of features which change in the presence of damage \cite{Farrar2012}. Determination of a baseline model can be either model driven, data driven, or a combination of both. As discussed with the example of guided waves in a fibre composite, analytical/numerical models can be difficult to develop and may not be robust enough to reasonably assume an accurate baseline state. Thus, there has been an increase in the use of purely data-driven models to determine a baseline state \cite{DaSilva2018}; however, it is then reasonable to assume this baseline state \textit{only} applies to nominally-identical scenarios/structures from which the data are collected. Since models generated from such methods are general and not specific to the scenario of interest, enhancing the data-driven models may be valuable in order to offer advantages, such as extrapolability, whilst maintaining accuracy from real data.

As mentioned above, this paper focusses on one guided wave feature, which is indicative of the amplitude of the first asymmetric mode. However, this does not limit the ideas, or even methods, of the work to just this feature; any energy-based features may be modelled with the same or a similar strategy, and the process of incorporating prior knowledge for other physics-based features can follow quite readily. For example, the time-of-flight modelling could be augmented by using prior knowledge of the dispersion characteristics of guided waves in the material.

\section{Experiment}
\label{sec:experiment}

\begin{figure}[h!]
    \centering
        \begin{subfigure}{.48\linewidth}
            \includegraphics[width=\textwidth]{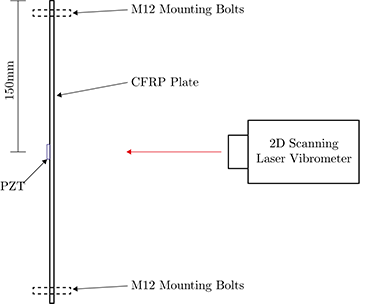}
            \caption{}
            \label{fig:PZT_loc}
        \end{subfigure}
        \begin{subfigure}{.48\linewidth}
            \includegraphics[width=\textwidth]{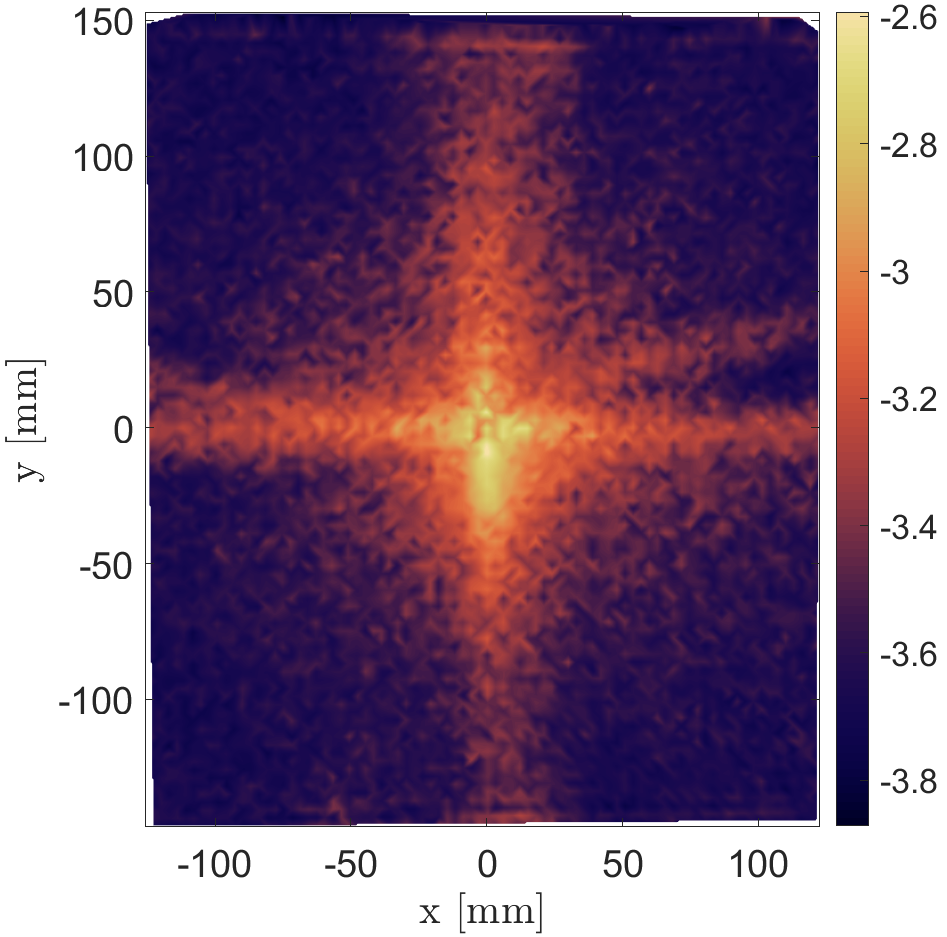}
            \caption{}
            \label{fig:ogScan}
        \end{subfigure}
    \caption{(a) Diagram showing a top down view of the experimental setup and location of PZT on the 300mm x 300mm CFRP plate and (b) results of feature-space map of $h_m$ over the surface of the plate from raw data, represented in $\log_{10}$ scale for viewing purposes.}
\end{figure}

Guided waves were initiated in a CFRP plate by excitation of a piezo-electric transducer (PZT), the location of which can be seen in \cref{fig:PZT_loc}. The PZT was actuated with a square pulse of temporal width $t_{pw}$, which would result in a frequency excitation range up to $1/t_{pw}$ Hz being excited, allowing multiple wave modes to be excited within the plate. A Polytec scanning laser vibrometer was used to measure the out-of-plane surface velocity of the induced wave-packets on the opposite side to the PZT, where the recording start was synchronised with the function generator attached to the PZT. Each wave-packet was then fed through a simple feature extraction process to generate a two-dimensional feature-space map of the maximum of the Hilbert envelope, $h_m$, over the surface of the plate. Specific details of the experimental setup are shown in \Cref{tab:exp_details}.

\begin{table}[h!]
    \centering
    \begin{tabular}{|l|l|}
        Plate dimensions & 300mm x 300mm x 1mm\\
        Layup & $[90/0/90]_s$, Epoxy matrix\\
        PZT Location & 150mm x 150mm \\
        Pulse width & 1$\mu$s\\
        Excitation bandwidth & 1MHz\\
        Signal record length & 4ms\\
        Pre-trigger & 400$\mu$s\\
        No. scan points & 8314\\
        No. of averages & 50\\
    \end{tabular}
    \label{tab:exp_details}
    \caption{Details of experimental setup used to acquire feature-space data.}
\end{table}

The results of the experiment showing the raw data feature-space map of $h_m$ over the surface of the plate can be seen in \cref{fig:ogScan}. One can clearly see the effect of the fibres on the amplitude of the first asymmetric mode; thanks to the phenomena of continuous mode conversion, the amplitude is greater when propagating along the fibres compared to when propagating across the fibres. This experimental procedure, though simple, is pivotal for the advanced tools, which will be presented later, in order to aid `black-box' data-based ML tools for development of a informed, data-driven (IDD) model. It is important to note the noisy data produced as a result of the low-signal-to-noise ratio that occurs from poor reflectivity of the material.

% \section{Feature space model generation}
\section{Structured machine learning tools}
\label{sec:models}

This section will discuss how, on the basis of physical understanding of the guided waves, data-based models of guided wave features in composites can be developed. %
Initially, a one-dimensional attenuation model is considered. %
This choice allows insight into the effect and contribution of the first two attenuation mechanisms described by Pollock\cite{Pollock1986}, geometric spreading and material damping. %
Following this investigation, two-dimensional Gaussian process models are considered. %
After considering the effectiveness of a `black-box' approach where the model is purely based on the data it has seen. %
The knowledge of the guided waves is included through two techniques; one of which is through incorporation of a mean function in the radial direction from the source. %
The mean function models the one-dimensional attenuation, i.e.\ it models the geometric spreading and material damping in the composite. %
The second approach is to include the physical knowledge of the guided waves through modifications to the kernel of the Gaussian process.

\subsection{One-dimensional attenuation modelling}

The one-dimensional model of the wave attenuation here is based on a Bayesian linear regression (BLR). %
A full description and some derivations for this method are provided by Murphy \cite{Murphy2012}, but a short introduction is given in \cref{app:BLR}. %
Several basis expansions of propagation distance $x$ were tested in combination in this work, the full model is shown in \cref{eq:BLR_basisComp}. %
These basis expansions correspond to different attenuation mechanisms associated with damping, geometric spreading and a combination of the two; these models are shown in \cref{tab:BLRmodelBasis}. %
The parameters ($ \Phi = \{\phi_1,\phi_2,\phi_3\} $) are switching parameters ($\phi_n=0,1$) which control the combination of functions shown in \cref{eq:BLR_basisLin}.
\begin{equation}
    A(x) = \beta_1^{\phi_1}\left(\exp(-\beta_2x)\right)^{\phi_2} \left(x^{-1/2}\right)^{\phi_3}
    \label{eq:BLR_basisComp}
\end{equation}
The model shown above is not a linear form that can be represented by \cref{eq:BLR_modelForm}, however, by taking the natural log of \cref{eq:BLR_basisComp} a \emph{linear-in-the-parameters} model can be developed. %
This is shown below in \cref{eq:BLR_basisLin}.

\begin{subequations}
    \begin{align}
    f(x) = \ln(A(x)) &= (\ln(\beta_1))\cdot\phi_1 - (\beta_2x)\cdot\phi_2 + (\ln(x^{-1/2}))\cdot\phi_3 \\
    \label{eq:BLR_basisLin}
    f(x) = \ln(A(x)) &= w_1\cdot\phi_1 - (w_2 x)\cdot\phi_2 + (\ln(x^{-1/2}))\cdot\phi_3 
    \end{align}
\end{subequations}

\noindent The values of $\beta_1$ and $\beta_2$ can be recovered by exponentiating $w_1$ and $w_2$ respectively.

\begin{table}[h!]
    \centering
    {\small
    \begin{tabular}{lll}
    \hline
      Model Basis & Linear Form & Basis Parameters \\
    \hline
      $ A_1(x) = \beta_1\exp(-\beta_2x) $        		 & $ f(x) = \ln(\beta_1) - \beta_2x $ & $ \Phi_1 = [1,\ 1,\ 0] $		\\
      $ A_2(x) = \beta_1x^{-1/2} $                 & $ f(x) = \ln(\beta_1) + \ln(x^{-1/2}) $ & $ \Phi_2 = [1,\ 0,\ 1] $		\\
      $ A_3(x) = \beta_1\exp(-\beta_2x) x^{-1/2} $     & $ f(x) = \ln(\beta_1) - \beta_2x + \ln(x^{-1/2}) $ & $ \Phi_3 = [1,\ 1,\ 1] $ 	\\
    \hline
    \end{tabular}
    }
    \caption{Bayesian linear regression model basis expansions}
    \label{tab:BLRmodelBasis}
\end{table}

\subsection{Two-dimensional attenuation modelling}
\label{sec:BLR}

So far, it has been shown how physical attenuation phenomena can be modelled as a Bayesian linear regression along one dimension. %
Now, attention turns to modelling the two-dimensional input feature space, which will be shown in \cref{sec:results} on a composite plate. %
If such a plate were homogeneous, modelling of the features along any one direction would provide an adequate model of the two-dimensional feature space. % 
However, in non-homogeneous materials --- such as composites --- this is no longer sufficient. %
Instead, the attenuation changes with direction, and therefore the model of the space must be able to capture changes in behaviour across the two dimensional field. %
For waves propagating from a point source, it can be helpful to think that there is a radial and angular component to the \emph{function} over the space which describes the feature of interest.

To build such a model it is necessary to have a tool which can model data across a two-dimensional space on the basis of observed data and which can be guided by belief about the physical phenomena. %
For this, a machine learning approach is adopted; %
the tool chosen for the job is a \emph{Gaussian process} (GP). %
The Gaussian process is a flexible Bayesian regression method which works by placing a prior over functions, which is then updated, on the basis of data, to return a posterior distribution over functions \cite{Rasmussen2005,o1978curve}. %
A brief introduction to Gaussian processes can be found in \cref{app:GP}. %

% \subsection{Kernel selection and design}

The kernel used in the GP is a significant modelling choice, and modifications of these provides structure through embedding prior belief of the model. %
These kernels are computed as any other kernel; linear pair-wise distances between points to form a covariance matrix. More detailed theory can be found in \cref{app:GP} and practical implementation can be found in \cite{Rasmussen2005}. %
There are a number of choices available for the kernel function, each of which embeds a different prior belief as to which \textit{family of functions} $f(\mathbf{x})$ is drawn from. %
For example, if a linear kernel is used, the solution to a Bayesian linear regression is recovered. More commonly, nonlinear kernels will be chosen, as many tasks require regression of nonlinear functions; %
popular choices include the use of the Squared-Exponential (SE) kernel or the Mat\'ern class of kernels. %
% These kernels, as well as being nonlinear, exhibit several other notable properties. %
% They are stationary, meaning they do not depend on the values of the two inputs, only the absolute distance between them. %
% Additionally, they are isotropic, i.e. they are insensitive to rotation, order, or translation of the inputs $\mathbf{x}$ and $\mathbf{x}'$ \cite{Rasmussen2005}. %
In most cases the mean function is set to zero in the prior; however, in this work it will be important to consider if the mean functional behaviour can be specified by the physics of the guided waves.

An important characteristic of GPs is that standard stationary kernels operate based on the Euclidean distance between two points, and so map covariances well when using a Cartesian space input. However, the physics and behaviour of guided waves is described here using the polar coordinate system, as they are emitted from a point source. Therefore, this attribute must be considered when utilising GPs for modelling the feature-space of guided waves. Padonou and Roustant \cite{Padonou2015} outline a method of applying GPs to a polar input space, where separate kernels are applied to the angular and radial dimensions separately, before combing to generate an overall covariance function.

\subsection{General nonlinear kernels}

It will be important to consider how the GP would model the data if no restrictions were placed on it with respect to the physical behaviour of the guided waves. %
This type of model will provide a benchmark against which the proposed models can be compared. %
Two important properties, which certain kernels possess, are stationarity and isotropy. %
A \emph{stationary} kernel is only dependent upon the difference between any two points, not the absolute values of those points. %
An \emph{isotropic} kernel is invariant to translation or rotation of the input data; practically, this appears as the covariance being only dependent on the absolute difference between two data points \cite{Rasmussen2005}. %
These properties will be important when discussing what is desired from a kernel to model the features of guided waves. %
 
One such  stationary and isotropic kernel, is the popular squared-exponential (SE) kernel \cite{Rasmussen2005}. %
This kernel is given by,
\begin{equation}
    k_{\textrm{SE}}(\mathbf{x},\mathbf{x}^\prime) = \sigma_f^2\exp\left\{-\frac{\vert\vert\mathbf{x}-\mathbf{x}^\prime\vert\vert_2^2}{2\ell^2}\right\}
    \label{eq:sqe_kernel}
\end{equation}
An alternative general nonlinear kernel is the Mat\'ern 5/2 kernel (as applied here to the radial dimension); this well-established kernel is used as it offers smooth shapes and is defined as,
\begin{equation}
    k_{\textrm{mat}}(\mathbf{x},\mathbf{x}^\prime) = \left(1+\frac{\sqrt{5}\vert\mathbf{x}-\mathbf{x}^\prime\vert}{\ell}+\frac{5\vert\mathbf{x}-\mathbf{x}^\prime\vert^2}{3l^2}\right) \exp\left(-\frac{\sqrt{5}\vert\mathbf{x}-\mathbf{x}^\prime\vert}{\ell}\right)
    \label{eq:matern_pol}
\end{equation}

% \noindent The SE kernel is controlled by two \emph{hyperparameters}, these are the signal variance $\sigma_f^2$, which controls the total scaling on the predicted function, and the length scale $\ell$, which controls the characteristic scale over which the function varies. %
% This notion can be more difficult to comprehend conceptually but can be interpreted by realising that a shorter length scale will lead to a greater change in the output for a smaller change in the input and vice versa. %
% Alternatively, it can be thought of as controlling the region of influence for each data point. %
% As the length scale increases, the function is affected by points further and further away from each other; this has the effect of smoothing through the data. %
% The estimation of these hyperparameters (and those in all the kernels) will be discussed in \cref{sec:hyperparameters}. %
% The SE kernel is infinitely differentiable so will produce smooth functions. %
% For engineering data collected from experimental studies this may not be a good assumption.

% So far, these kernels have been stated for a general input vector $\mathbf{x}$. %
The simplest and perhaps most obvious choice for mapping the features across a composite plate would be to set $\mathbf{x}=\left\{x,y\right\}$, the cartesian coordinates of a location on the plate. %
% Applying this with a squared exponential kernel will form the baseline model for this work. %
This method imposes the prior belief that the feature being modelled across the plate will vary smoothly in a nonlinear manner with respect to the $x$ and $y$ coordinate. %
% In many senses this is the simplest model that could be chosen, it is also the most flexible as it imposes very little restriction on the form of the functions that can be modelled.

However, since it is known here that the waves are generated from a point source in the centre of the plate and that these will propagate from that point, the behaviour would be better modelled in a set of polar coordinates. %
For the general case, here the work of Padonou and Roustant \cite{Padonou2015} is followed, and a detailed description is provided in \cref{app:polarGP}. %
An important distinction of the polar kernel is the definition of distances between points, specifically in the angular dimension where, for example, a point with angle $359^\circ$ should have a high covariance with $1^\circ$ if the radii are also close. %
The general polar kernel is defined as,
\begin{equation}
	k_2(\mathbf{x},\mathbf{x}') = \sigma_f^2\left(1+\sigma_{f,r}^2k_{\textrm{mat}}(\rho,\rho')\right)\left(1+\sigma_{f,a}^2k_{\textrm{W}}(\theta,\theta ')\right)
	\label{eq:polar_kern}
\end{equation}
\noindent where $\sigma_{f,m}$ and $\sigma_{f,a}$ act as weights representing the influence of changes in each dimension on a change in the output.

\begin{figure}[h!]
    \centering
    \includegraphics[width=0.85\textwidth]{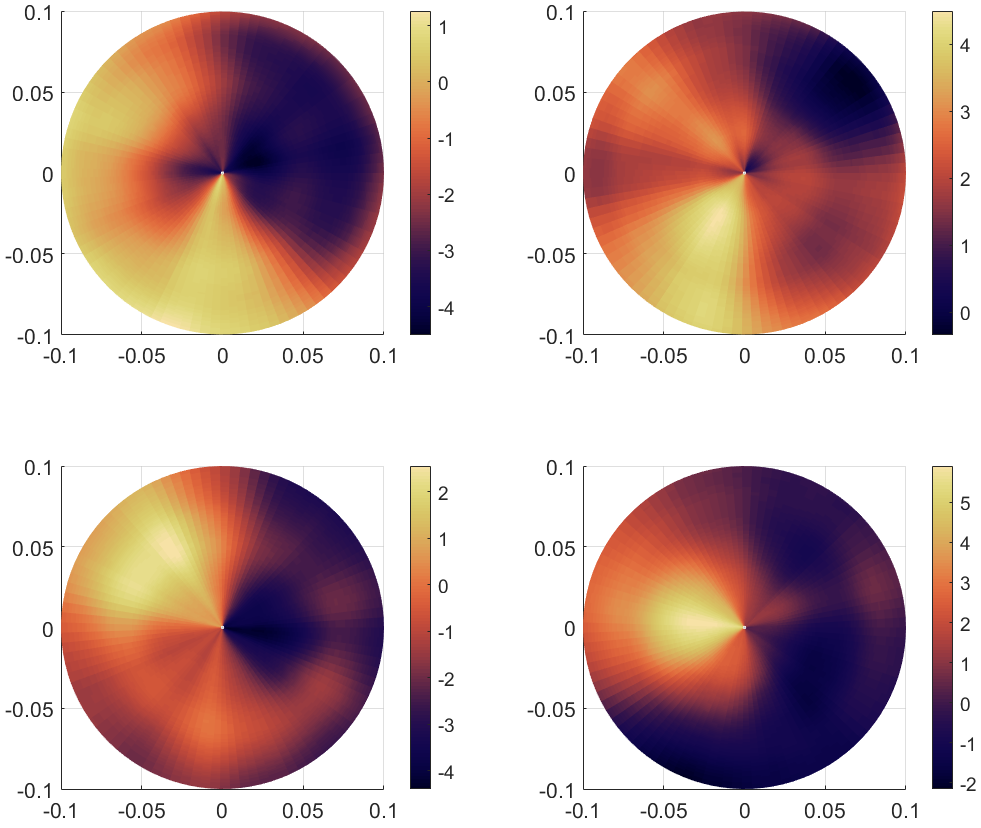}
    \caption{Selection of four random polar space functions randomly generated from covariance function represented in \cref{eq:polar_kern}, with hyperparameter values of $\tau=4$, $l=1$ and $\sigma_{f,m}^2=\sigma_{f,a}^2=1$.}
    \label{fig:2D_priors_Pad}
\end{figure}

Since the GP is a generative model over functions, it is possible to sample realisations of possible functions from the distribution, allowing the user to understand the type of functions that the kernel will generate. %
% If this is done without conditioning the model on any data points, it is possible to see what realisations from the \emph{prior} may look like. %
% This strategy can be helpful as it allows the user to understand heuristically the type of functions that the kernel they have chosen will generate. %
% In this case, it will be used to show how the models can generate behaviour which appears closer to what would be expected physically \emph{a priori}.
Using this method, four prior realisations were generated from the polar kernel shown in \cref{eq:polar_kern} and can be seen in \cref{fig:2D_priors_Pad}. %
It can be seen that the functions generated appear to operate separately on each dimension $\rho$ and $\theta$. %
A key characteristic to note, is that there is no discontinuity as $\theta$ moves through $2\pi$ to zero in the angular direction; this is as a result of the Wendland-$C^2$ function kernel (\cref{eq:wend_def,eq:cov_ang}). %
% Because of this a general nonlinear kernel has been developed for functions over a polar coordinate space. %
Further discussions on the characteristics of such a kernel can be found in \cite{Padonou2015}. % 
For this work, the polar kernel will serve as an alternate model where there is very little restriction placed on the functions that can be modelled. %
% It can also be considered not to embed any knowledge of the physics of the process except that the function can be well represented in a polar coordinate space.

\subsection{Mean function modelling of attenuation}

% So far, the models shown can be considered to embed very little prior knowledge about the process. %
% The restrictions placed on the function being modelled are only that it is nonlinear, stationary, smooth (as determined by choice of Mat\'ern function), and that it may be best modelled in a polar coordinate system. %
% However, given the wealth of literature on propagation of guided waves, it would seem sensible to attempt to make use of this in any modelling procedure.

From this point onwards, the model learning will make use of prior knowledge of guided wave propagation. %
This informed model learning will begin by considering how a mean function $m(X)$ can be used to introduce a physical basis to the model. %
There is no restriction on this mean function given that it is known. %, it could be, for example, a polynomial or Fourier expansion. %
Mathematically, it is trivial to include the mean function (if known) through simply subtracting the expected mean function from the target data and training on the residuals,

\begin{equation}
    \tilde{\mathbf{y}}=\mathbf{y}-m(\mathbf{x})
\end{equation}

This can be interpreted as learning the difference or discrepancy between this chosen mean function and the generating function of the data. %
In this scenario, the mean function $m(\mathbf{x}) = \phi(\mathbf{x})\mathbf{w}$ is the model described in \cref{sec:BLR}, the one-dimensional Bayesian linear regression model. %
Since the weights of the model vary depending on propagation direction with respect to fibre orientation, it is necessary to simultaneously learn the distribution of $\mathbf{w}$ the weights of the BLR and the hyperparameters of the GP. %
Therefore, the linearised form shown in \cref{eq:BLR_basisLin} is used and the target data becomes,
\begin{equation}
    \tilde{\mathbf{y}}=\ln(\mathbf{y})-\ln(\rho^{-1/2})-m(\mathbf{x})
\end{equation}
and the steps for training and expected values of the mean and variance can be followed in \cite{Rasmussen2005}. %
This solution can be interpreted as finding the mean one dimensional behaviour across all propagation directions. %
This mean behaviour is then compensated by the GP to fit the observed data and learn the latent function which describes it.
% This stage will form the first model considered in the paper which attempts to incorporate understanding of the physical processes involved in the propagation of guided waves into the machine learning methodology.

\subsection{Functional priors through kernel design}

An alternative to using the mean function to include what is known about guided waves, is to modify the kernel. %
% Remembering that the kernel embeds belief as to the \emph{family of functions} from which the data is generated, 
It is possible to restrict the \emph{family of functions} \emph{a priori} to generate only functions which are plausible, given physical understanding of the guided waves. %
This restriction is the key advantage of the proposed Bayesian approaches in this paper. %; that knowledge about the functional process can be embedded in a formal and rigorous manner without overly restricting the model's ability to learn from data.

All of the models which are generated from this point onwards will consider the propagation of the guided waves to occur in a polar coordinate system, where the source is located at radius $\rho=0$, i.e.\ the source is at the origin. %
% These kernels will build upon the work of \cite{Padonou2015}, by also incorporating prior belief about the physical form of the guided-wave feature space.
Since the kernel which defines the process will be composed of the ANOVA combination of the radial component and the angular component, it is possible to consider how to modify each of these components individually. %
% In other words, it will be shown how expected radial and angular behaviour can be embedded in isolation from the other. %
% This separation of dimensions in modelling is especially useful, since varying amounts of information may be available for each of these.

\subsubsection{Imposing rotational symmetry in the feature space}
% Initially, the angular component of the kernel is considered. %
It can be seen in \cref{fig:ogScan}, that the energy of the wave exhibits a symmetry on the plate. %
Physically, this makes sense given what is known about the symmetry in the orientation of the fibres in the lay-up; it is therefore desirable to exploit this in the kernel. %
\Cref{eq:sym_kernel} was designed to model this symmetric behaviour. %
The strictly periodic kernel is applied to the angular dimension, where $n$ can be altered to include the number of axes of symmetry,
\begin{equation}
    k_{\textrm{sym}}(\theta,\theta ') = \left(\alpha_1 + \alpha_2\cos(2n d_2)  \right), \qquad n\geq 1
    \label{eq:sym_kernel}
\end{equation}

\noindent $d_2=\arccos(\cos(\theta-\theta '))$ is the geodesic distance, $n$ the number of symmetry axes required, $\alpha_1$ is the offset term, and $\alpha_2$ the amplitude hyperparameter. %

\begin{figure}[h!]
    \centering
    \includegraphics[width=\textwidth]{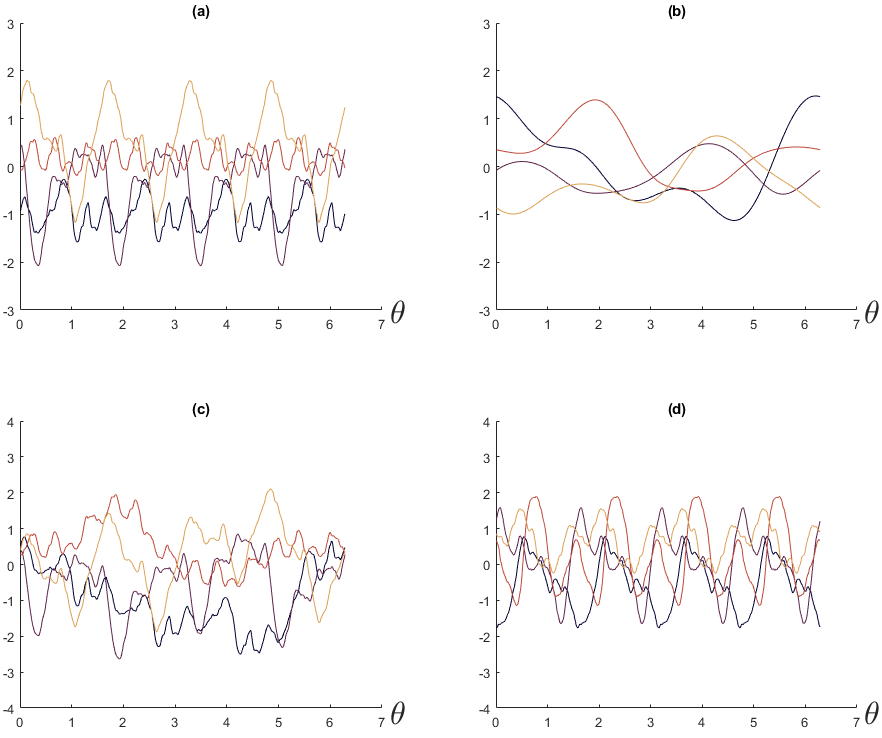}
    \caption{Selection of random priors for the angular kernel designs, over a full circle range, for (a) strictly-periodic kernel (\cref{eq:sym_kernel}), (b) squared-exponential kernel (\cref{eq:sqe_kernel}), (c) multiplied combination and (d) additive combination of the kernels (\cref{eq:ang_kernel}). Each line represents a different random function drawn from these priors.}
    \label{fig:ang_priors}
\end{figure}

Sample functions from this prior are shown in \cref{fig:ang_priors}a. %
These samples show the strict periodicity that this kernel imposes. %
Notably, this form of kernel does not enforce the phase of the function. %, i.e.\ the function is not necessarily a maximum or zero when $\theta=0^{\circ}$. %
Instead, as data are observed, the posterior distribution --- or function that is learnt --- adapts to the phase information in the data.  %
This adaptability is a benefit since enforcing phase within the kernel may lead to issues if the fibre orientation is not known exactly. %

The primary issue with the strictly periodic kernel is the assumption of consistent rate of variation in the function with $\theta$, which may make \cref{eq:sym_kernel} too restrictive to model accurately the guided wave feature space. %
It is clear to see in \cref{fig:ogScan}, that the energy of the wave decays rapidly in the transition between propagation along fibres and across fibres as the angle $\theta$ varies. %
% This could be corrected in two ways; the first could apply a variable length scale in \cref{eq:sym_kernel} as a function of the angle $\theta$. %
% This method is difficult since it would be necessary to define this variation \emph{a priori}, and it is not known exactly. %
Therefore, flexibility was introduced by an additive combination of the strictly-periodic kernel and the squared-exponential kernel previously shown in \cref{eq:sqe_kernel}.

% As discussed, the squared-exponential kernel generates smooth nonlinear functions; this can be seen in the prior draws shown in \cref{fig:ang_priors}b. %
% The squared-exponential kernel was used due its characteristic smoothness, as can be seen in \cref{fig:ang_priors}b, which can model deviations from the angular kernel. %
As well as allowing varying rate of change with $\theta$, a combined kernel also reduces some of the restrictions that are imposed with the pure strictly-periodic kernel \cref{eq:sym_kernel}. %
An additive combination was used as opposed to multiplicative, as this does not generate large variations in amplitude between period peaks and allows for the kernel to capture the symmetry, while still allowing some variation to take place. %
The additive combination can be considered an `OR' operation \cite{duvenaud2014automatic}, the resulting kernel applied to the angular dimension is given by,
\begin{equation}
    k_{\textrm{ang}}(\mathbf{\theta},\mathbf{\theta}') = \sigma_{f,\textrm{sqe}}^2 \exp\left(-\frac{d_2^2}{l_1^2}\right) + \sigma_{f,\textrm{sym}}^2\left(\alpha_1+\alpha_2 \cos\left(nd_2\right)\right)
    \label{eq:ang_kernel}
\end{equation}
An important point of the resulting kernel is that it is stationary, as it is only proportional to the distance between points, rather than their values. %
This means that the kernel is unaffected by translation or rotation of the coordinates, a key advantage when modelling in the angular dimension.

\subsubsection{Kernel forms for radial attenuation}

% From physical knowledge of the attenuation of guided waves, kernels can be structured to emulate the effects of propagation distance, $\rho$, on the energy of the wave. %
As discussed in Section \ref{sec:UGW_physics}, two known and documented attenuation mechanisms can be modelled; these are shown in \cref{eq:Lamb_atten_geo,eq:Lamb_atten_damp}. %
Attenuation effects due to viscoelastic damping of a material can be embedded into the priors through the use of an \textit{exponential decay} (ED) kernel, where propagation distance $x$ is replaced with $\rho$,
\begin{equation}
    k_{\textrm{ed}}(\mathbf{\rho},\mathbf{\rho}') = \exp\left(-\mathbf{\rho}l\right)\cdot\exp\left(-\mathbf{\rho}'^{\top}l\right)
    \label{eq:expDec_kernel}
\end{equation}
Attenuation effects due to geometric spreading can be modelled through the use of a polynomial kernel (Eq. \ref{eq:poly_kernel}), where $p=-1/2$, in order to represent Eq. \ref{eq:Lamb_atten_geo}.
\begin{equation}
    k_{\textrm{pol}}(\mathbf{\rho},\mathbf{\rho}') = (\mathbf{\rho}\cdot\mathbf{\rho}'^{\top})^p
    \label{eq:poly_kernel}
\end{equation}

% As such, independent kernels have been developed which account for two of the physical attenuation mechanisms; these being the exponential decay due to viscoelastic damping and the power law related to geometric spreading. %

\begin{figure}[h!]
    \centering
    \includegraphics[width=\textwidth]{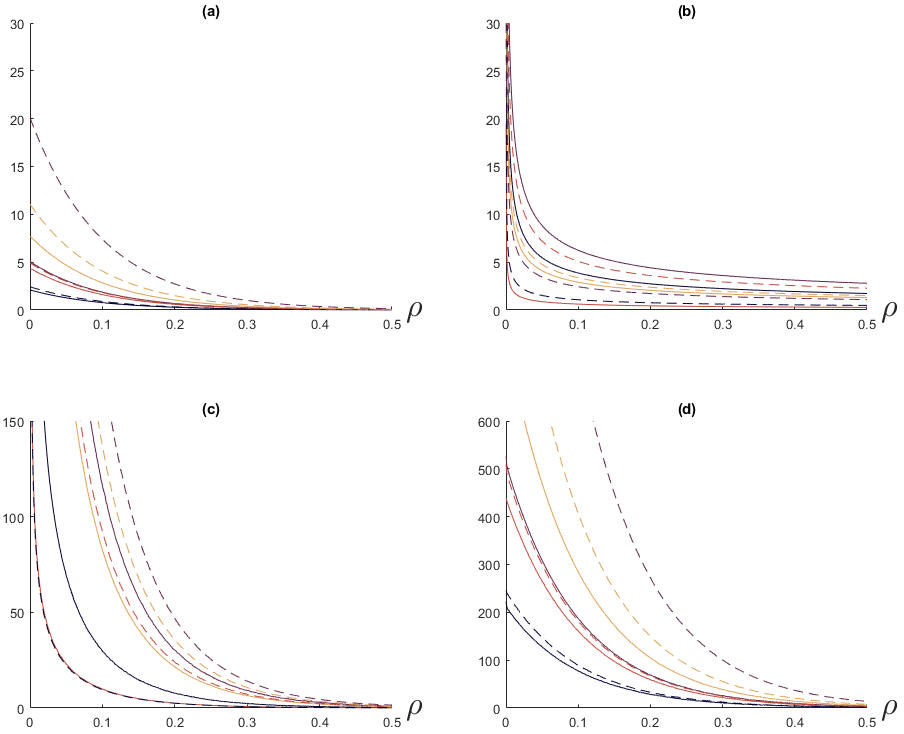}
    \caption{Selection of random priors for the radial kernel designs, for (a) exponential decay kernel (\cref{eq:expDec_kernel}), (b) square-root decay (\cref{eq:poly_kernel}), (c) multiplicative combination (\cref{eq:rad_kernel}) and (d) additive combination of the kernels. Each line represents a different random function drawn from these priors.}
    \label{fig:rad_priors}
\end{figure}

Prior draws from both the ED and polynomial kernels can be seen in Figs. \ref{fig:rad_priors}(a) and \ref{fig:rad_priors}(b) respectively. %
Both kernel functions embed decay with respect to propagation distance $\rho$, but each model shows a different mechanism for this decay. %
% Importantly, the power law controlling the decay due to geometric spreading is considered known, so is specified as $p=-1/2$. %
The kernel chosen to represent decay due to the geometric spreading will always tend to infinity as $\rho\rightarrow0$ and this limitation should be considered. %
% The ED kernel also models a decay behaviour as the radial distance from the source increases; it is expected that its effect will be much less than the effect of geometric spreading, but will still be important to capture the behaviour of the guided waves.

To illustrate the use of these kernels, multiplicative and additive combination of these two attenuation mechanism kernels can be seen in Figs. \ref{fig:rad_priors}(c) and \ref{fig:rad_priors}(d) respectively. %
% Reference \cite{Schubert2011} discusses accounting for geometric attenuation by adding the $\sqrt{x_0/x}$ into \cref{eq:Lamb_atten_damp} and replacing $x$ with $x-x_0$. %
A multiplicative combination of the attenuation mechanisms aligns more closely with physical understanding and the discussion presented in \cite{Schubert2011}, since the kernels operate to reduce the energy in the wave simultaneously and do not subtract energy, but rather reduce it. %
As such, a kernel to model attenuation along the radial direction is proposed as the multiplicative combination of the exponential decay (\cref{eq:expDec_kernel}) and polynomial (\cref{eq:poly_kernel}) kernels; this is
\begin{equation}
    k_{\textrm{rad}}(\mathbf{\rho},\mathbf{\rho}') =  \sigma_{f,r}^2 (\mathbf{\rho}\cdot\mathbf{\rho}'^{\top})^p \; \cdot \; \left(\exp\left(-\mathbf{\rho}l_2\right)\cdot\exp\left(-\mathbf{\rho}'^{\top}l_2\right) \right)
    \label{eq:rad_kernel}
\end{equation}

\subsubsection{Combined two-dimensional kernel}
\label{sec:combKern}

It has been shown how understanding of the physical processes involved in attenuation of guided waves can be used to impose prior belief in the GP machine learning model along each of the radial and angular dimensions. %
It remains to explain how these may be combined to form a meaningful prior over the two-dimensional feature space. %

Following closely the approach of \cite{Padonou2015} for the general nonlinear kernel in polar coordinates, the two kernels described in \cref{eq:ang_kernel,eq:rad_kernel} will be combined using an \textit{ANOVA} approach,
\begin{equation}
    k_3(\mathbf{x},\mathbf{x}') = \left(1+\sigma_{f,a}^2k_{\textrm{ang}}(\mathbf{\theta},\mathbf{\theta}')\right)\left(1 + \sigma_{f,r}^2k_{\textrm{rad}}(\mathbf{\rho},\mathbf{\rho}')\right)
    \label{eq:UGW_kernel}
\end{equation}
where $\mathbf{x} = \{ \{\theta,\rho\}_1, ... , \{\theta,\rho\}_m \}$. %

\begin{figure}[h!]
    \centering
    \includegraphics[width=\textwidth]{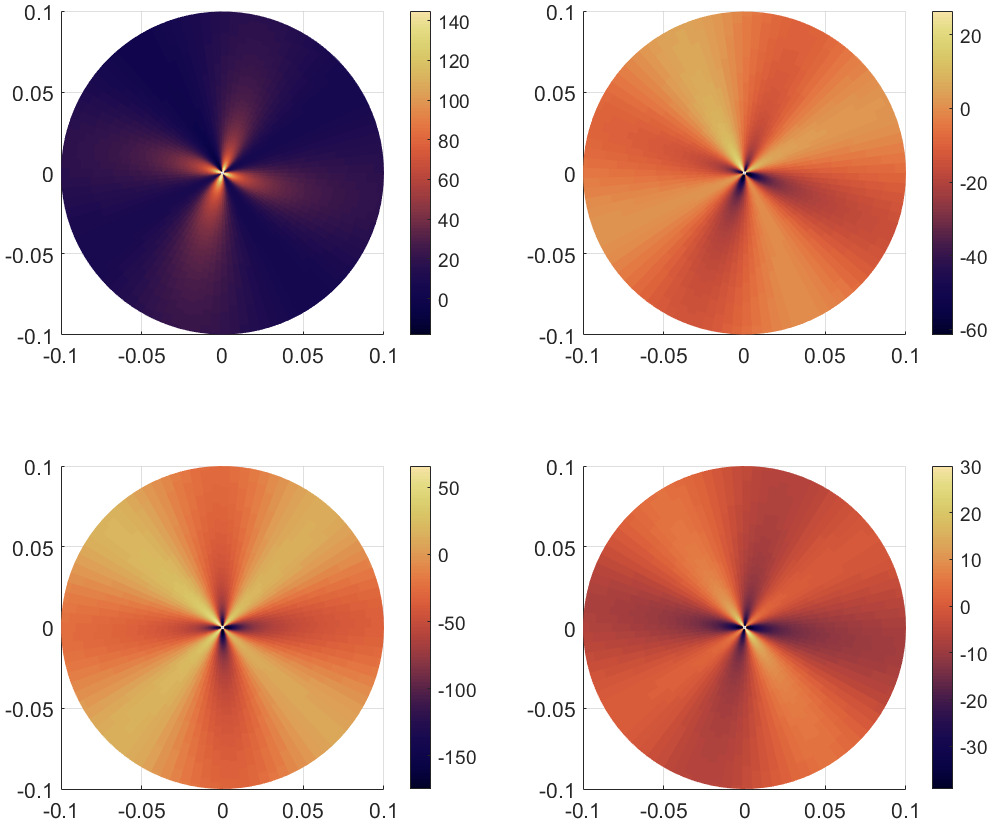}
    \caption{Selection of four random polar space functions, selected from 1000 random functions generated from covariance function represented in \cref{eq:UGW_kernel}, with hyperparameter values of $\mathbf{\Theta}_k=\{1,10,10,1,1,1,1,1,0.001\}$. Here the functions are non-dimensional as the functions are samples from an arbitrary prior.}
    \label{fig:2D_priors_UGW}
\end{figure}

Again, it is possible to draw samples of the functions, now in the two-dimensional space, to visualise the restrictions which have been placed on the functions that can be modelled. %
Four prior draws from a GP with zero mean and the covariance defined in \cref{eq:UGW_kernel} can be seen in \cref{fig:2D_priors_UGW}. %
It should be noted at this point that the input units and output values are non-dimensional, and the figures showcase key characteristics imposed by the kernels by displaying functions that are samples from an arbitrary prior. %
It is reassuring that these prior draws match, at least visually, the behaviour that would be expected in the data being modelled. %
This type of \emph{prior predictive checking} can be invaluable for confirming that the assumptions built into the model are reasonable. %
A key feature that can be seen is the symmetry that this introduced in the angular dimension without requiring a fixed phase to be specified. %
It can also be seen that slight variations from this symmetry are possible due to the inclusion of the squared-exponential kernel in \cref{eq:ang_kernel}. %
The decay in the radial direction as a result of the kernel shown in \cref{eq:rad_kernel} is also clearly seen. %
As a result of these characteristics it has been shown how a GP kernel can be designed in such a way that it is applicable to modelling the attenuation of guided waves in a two-dimensional space. %

A second kernel was also tested with the same radial component as described in \cref{eq:rad_kernel}, but with an alternative angular kernel. %
For this method, $k_{\textrm{ang}}(\theta,\theta')$ becomes a modified version of \cref{eq:cov_ang}, where the geodesic distance is instead defined as $ d_2(\theta,\theta ') = \arccos(\cos(2n(\theta-\theta '))) $, where $n$ is again the number of symmetry lines required. %
This alteration was done to still enforce symmetry but allow a more flexible modelling of the functions being considered in the angular dimension. %
This kernel has the form,

\begin{equation}
    k_4(\mathbf{x},\mathbf{x}') = \left(1+\sigma_{f,a}^2k_{\textrm{W}}(\mathbf{\theta},\mathbf{\theta}')\right)\left(1 + \sigma_{f,r}^2k_{\textrm{rad}}(\mathbf{\rho},\mathbf{\rho}')\right)
    \label{eq:UGW_kernel_k4}
\end{equation}

\begin{figure}
    \centering 
    \includegraphics[width=\textwidth]{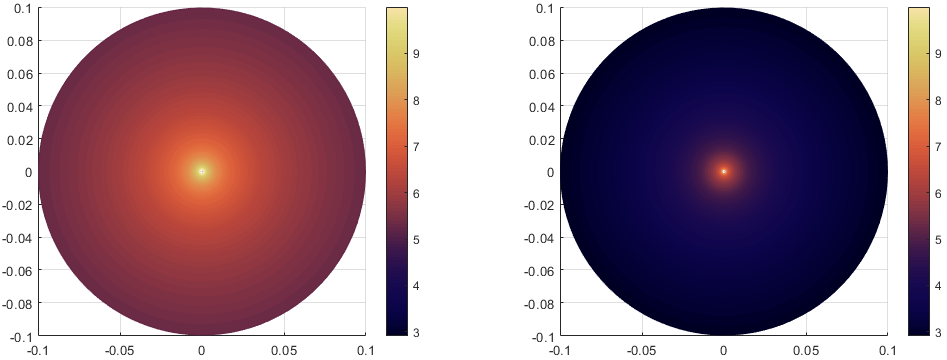}
    \caption{Prior variances over polar coordinate space from kernel E (left) and kernel F (right), represented by \cref{eq:UGW_kernel,eq:UGW_kernel_k4} respectively. Variances are plotted in $\log_{10}$-scale for viewing purposes.}
    \label{fig:2D_priorVar_k3k4}
\end{figure}

So far, only samples from each GP have been shown. %
However, it is possible to recover the distribution over the model in closed form. %
Since the mean function chosen in most models is simply zero across the complete space, the prior mean is not very informative. %
The prior variance, however, is of interest to consider. %
In \cref{fig:2D_priorVar_k3k4} the prior variances of \cref{eq:UGW_kernel,eq:UGW_kernel_k4} are shown. %
It is important to remember that this is the variance in predictions made by the model before the information from any data has been included.
The variance for both of these kernels is seen to decay as the distance from the source increases. %
The model will tend towards infinity at $\rho=0$ for two reasons; the kernels are non stationary \cite{Rasmussen2005}, and due to the exponential decay included through \cref{eq:expDec_kernel}, the function values themselves will tend towards infinity as seen in \cref{fig:2D_priors_UGW}. %
This limitation of the models should be considered and care should be taken if predicting close to $\rho=0$. %
However, in the experimental data used in this study the waves are generated by means of a piezoelectric actuator. %
This means that the source of the guided waves is not a point source and the models should not be used within the region covered by the piezoelectric actuator. % 

\subsection{Overview of modelling approaches}

Up to this point, it has been discussed how one might construct a GP kernel which can represent the behaviour of guided wave attenuation. %
It is worth reviewing the models which will be compared when results are shown on experimental data. %
\Cref{tab:kernModels} shows a summary of all six models which will be compared in this study. 

As a baseline, model A is the archetypal Gaussian process model with a zero-mean function and the use of the squared-exponential kernel operating on two inputs, the $x$ and $y$ coordinates on the plate. % 
This provides a benchmark where no knowledge of the guided-waves is included. %

The second model (B) is a demonstration of the use of the polar coordinate GP of \cite{Padonou2015}. %
This model also contains no specific reference to the physical mechanisms in guided-waves, but does make use of the knowledge that the guided-waves propagate radially from a source. %
The use of a polar coordinate system in this case is a sensible choice given the structure of the data being used. %
This method serves as another benchmark demonstrating an approach which requires very little understanding of the physical mechanisms involved in guided wave propagation.

Model C is the first model where a specific physical process is included. %
In this case, the model of wave attenuation $A_3(x)$ (\cref{tab:BLRmodelBasis}) is used along the radial direction as the mean function. %
The kernel used is the same flexible polar kernel as in model B --- that proposed in \cite{Padonou2015}. %
Importantly, this kernel remains flexible to influence the model in both the radial and the angular dimensions, potentially correcting for any unmodeled phenomena along the radial dimension in the mean function. %

Model D restricts the flexibility of Model C by removing the dependence of the kernel on the radial dimension. %
The GP used here relies on the mean function to capture all of the radial behaviour through $A_3(x)$ and the covariance to capture all the variation in the angular dimension. %
This implies that the data can be generated by some function, offset from the mean, which is only dependent upon the angle being considered. %
This model should be considered with care since it is highly restrictive. %

Model E removes the use of the mean function; instead the knowledge of the guided waves is embedded directly in the kernel as a \emph{functional prior}. %
This model enforces periodicity in the angular dimension and embeds the physical attenuation models in the radial direction by means of the kernel described in \cref{eq:UGW_kernel}. %
Model F is very similar to the model E but with the kernel defined as in \cref{eq:UGW_kernel_k4}, with the modification to the angular component described previously.

\begin{table}[h!]
    \centering
    \begin{tabular}{ c | c | c | c }
        Model & Mean $m(\mathbf{x})$ & Covariance $k(\mathbf{x},\mathbf{x}')$ & Input space \\
         \hline
         \hline
         A & 0 & $k_{\textrm{SQE}}(\mathbf{x},\mathbf{x}')$ & Cartesian \\
         B & 0 & $k_{2}(\mathbf{x},\mathbf{x}')$ & Polar \\
         C & $A_3(\rho)$ & $k_{2}(\mathbf{x},\mathbf{x}')$ & Polar \\
         D & $A_3(\rho)$ & $k_{\textrm{W}}(\theta,\theta')$ & Polar \\
         E & 0 & $k_3(\mathbf{x},\mathbf{x}')$ & Polar \\
         F & 0 & $k_4(\mathbf{x},\mathbf{x}')$ & Polar \\
    \end{tabular}
    \caption{Table of GP strategies tested for feature-space mapping, showing the properties and characteristics of each model.}
    \label{tab:kernModels}
\end{table}

\cref{app:kernels} provides a reference in which each of these kernel forms can be compared and in which the hyperparameters are listed. %
The reader may find this a useful companion if planning to reproduce the methodology from this work.

\subsection{Hyperparameter learning}
\label{sec:hyperparameters}

Thus far, the kernels of the GP have been presented as priors over the functions which that GP will generate. %
These modelling strategies have allowed embedding of physical processes governing the behaviour of guided waves in a flexible and rigorous manner. %
However, each of these kernels has a small number of associated \emph{hyperparameters} which govern the characteristics of the family of functions they represent.
It is necessary, therefore, to review how a user may practically ascertain the values of these hyperparameters. %
As with many problems in machine learning, and indeed engineering, this boils down to an optimisation problem. %
The specific form of this problem will now be shown. 

The hyperparameters vary depending upon the form of the kernel, but for generality $\mathbf{\Theta}_k$ is considered to be the vector of hyperparameters for whichever kernel is being used.
For example, in the case of the kernel proposed by \cite{Padonou2015}, this vector is defined as $\mathbf{\Theta}_k=\{l,\sigma_{f,r}^2,\sigma_{f,a}^2,\tau\}$; whereas, for the kernel in \cref{eq:UGW_kernel} which is used in model E, this vector is $\mathbf{\Theta}_k=\{l_1,\alpha_1,\alpha_2,\sigma_{f,\textrm{sqe}}^2,\sigma_{f,\textrm{sym}}^2,\allowbreak l_2,\sigma_{f,r}^2,\sigma_{f,a}^2,\sigma_n^2\}$. %
The hyperparameters each control distinct and important characteristics to the kernel.
As an example for the kernel used in model E (\cref{eq:UGW_kernel})), these characteristics are interpreted as follows:
\begin{enumerate}
    \item $l_1$ and $l_2$ are the characteristic length scales of the SE kernel and ED kernel respectively.
    \item $\alpha_1$ and $\alpha_2$ represent the offset and scaling term for the sine-wave-based strictly-periodic kernel.
    \item $\sigma_{f}^2$ terms are scaling factors for individual kernels which control their relative importance when combined.
    \item $\sigma_n^2$ is the noise variance parameter related to the expected measurement noise on the signal.
\end{enumerate}

In this work, the parameters of the mean function are also included in the optimisation routine to estimate them synchronously with the hyperparameters of the kernel.
This optimisation problem is most commonly cast as the maximisation of the \textit{marginal likelihood} of the Gaussian process which is available in closed form \cite{Rasmussen2005}. %
This optimisation strategy has several advantages such as leveraging the Bayesian Occam's Razor \cite{rasmussen2001occam}. %
For computational reasons this method is realised practically as a minimisation of the \textit{negative log marginal likelihood} (NLML), more details of which can be found in \cite{Rasmussen2005} and \cref{app:GP}. %

As such, the optimisation task at hand is formally to estimate,
\begin{equation}
	\hat{\mathbf{\Theta}}= \textrm{arg min}(-\log p(\mathbf{y}\vert\mathbf{\Theta}))
	\label{eq:min_nlml}
\end{equation}

For all results in this paper, \cref{eq:min_nlml} was optimised using the \textit{quantum particle swarm} technique \cite{Yang2004}. %
Discussion of this choice can be found in \cite{rogers2017choice}.

\section{Results}
\label{sec:results}

Having developed a number of approaches for modelling of guided-wave features, it is necessary to demonstrate the differences between each of these on an experimental dataset. %
The dataset is chosen is the composite plate described in \cref{sec:experiment}. %
Initially, it will be shown how the one-dimensional models described in \cref{sec:BLR} can be applied along a fixed radial direction. %
Following this, the two-dimensional models laid out in \cref{tab:kernModels} will also be compared.

\subsection{One-dimensional attenuation models}

\begin{figure}[h!]
    \begin{subfigure}{0.48\textwidth}
        \includegraphics[width=\textwidth]{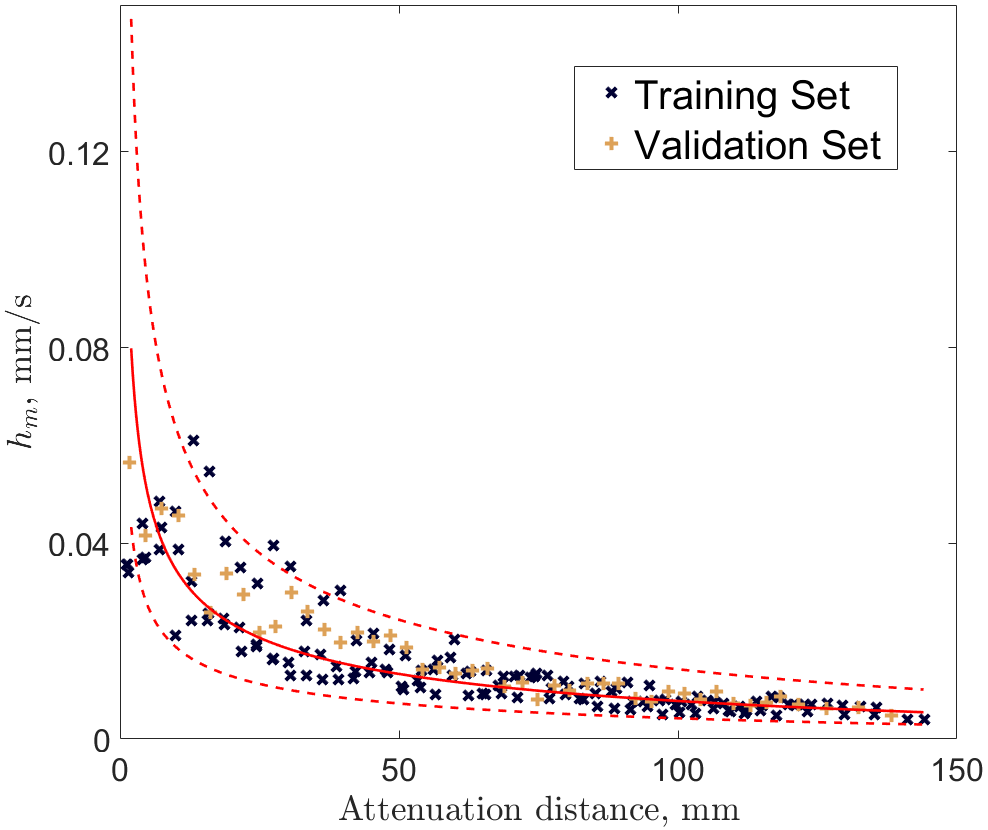}
        \caption{}
        \label{fig:phi3_dir}
    \end{subfigure}
    \begin{subfigure}{0.48\textwidth}
        \includegraphics[width=\textwidth]{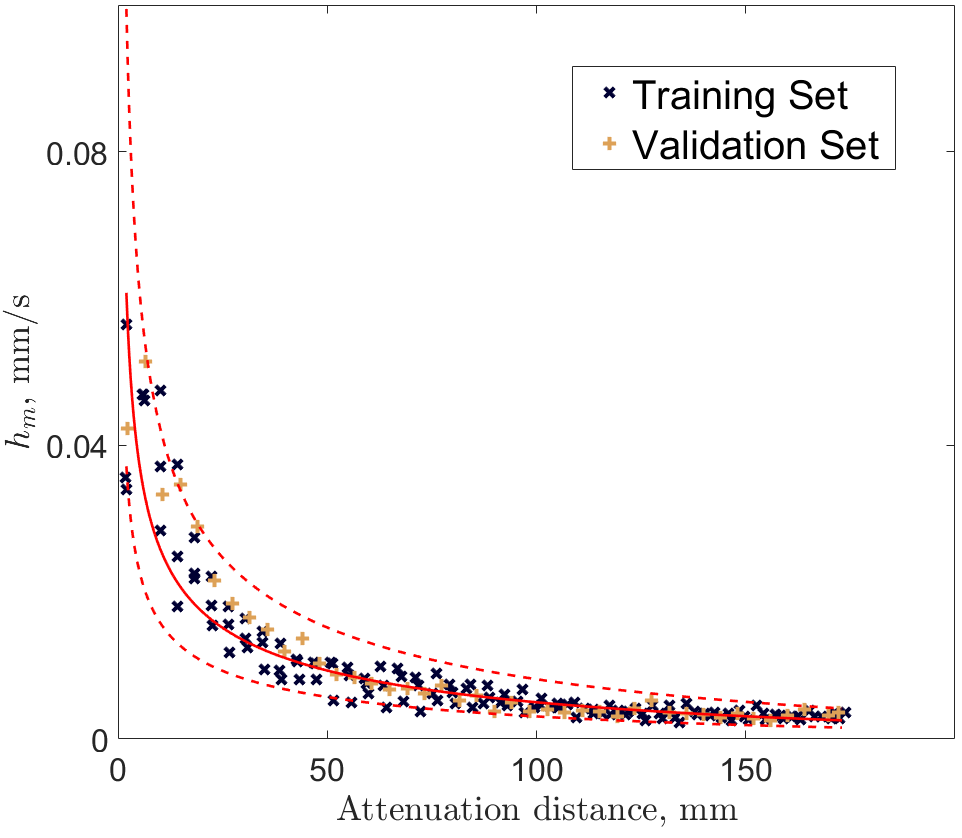}
        \caption{}
        \label{fig:phi3_indir}
    \end{subfigure}
    \caption{Models of attenuation of the energy of the first antisymmetric mode using model $\Phi_3$ for propagation directions (a) along the fibres and (b) through the fibres. The estimated mean function is plotted with a solid red line and the 95\% confidence interval is shown with dashed red lines.}
    \label{fig:BLR_results}
\end{figure}

The models shown in \cref{sec:BLR} are compared for two different cases; the first is where data are collected along the fibres in the weave and the second when across the fibres. %
The $NMSE$ is calculated for both the training set and the test data sets. %
\Cref{fig:BLR_results} and \Cref{tab:BLR_results} show the results of this process. %
From the values of the $NMSE_{tr}$ and $NMSE_{t}$ for both propagation orientations, it appears that the multiplicative combination of geometric spreading and material damping is the most promising model for the attenuation of the energy of the first antisymmetric mode for all propagation directions in a CFRP plate. %
It is also interesting to note the increased quality of fit of model $\Phi_2$ in comparison to $\Phi_1$ indicates that geometric spreading has a larger effect on the attenuation of the energy than material damping.

When modelling the attenuation of the guided-waves propagating through the fibres, the 95\% confidence intervals, seen in \cref{fig:BLR_results}, are much smaller, which is due to the less noisy distribution of values for $h_m$ around the estimated mean. %
Physically, this could be envisaged as the increased number of boundaries for the wave to propagate through causing more frequent mode conversion and so there is a more consistent energy dissipation as it tends towards \textit{continuous mode conversion} \cite{Mook2014,Willberg2012}. In contrast, when propagating along the fibres, the wave mode is relatively uninterrupted in comparison, and so its initial energy has a strong effect on the resulting shape. %
Furthermore, attenuation due to geometric spreading is more likely to be described with \cref{eq:Lamb_atten_geo} during CMC. %
However, this is currently an assumption, experimental validation of this is planned as future work.%

\begin{table}[h!]
    \centering
    \begin{tabular}{ l | l | l | l | l | l | l}
         & $\beta_1$  & $\beta_2$  & $\sigma^2$ & $\sigma_{\mathbf{w}}$ & $NMSE_{tr}$    & $NMSE_{t}$\\
         \hline
         \multicolumn{7}{c}{Along fibres}                                        \\
         \hline
    $\Phi_1$ & 0.001523 & 0.01215   & 9.403  & 0.1532  & 0.13203 & 0.10548 \\
    $\Phi_2$ & 0.004844 &           & 8.199  & 0.0640  & 0.21864 & 0.11664 \\
    $\Phi_3$ & 0.005934 & 0.00288   & 6.558  & 0.1068  & 0.17489 & 0.14795 \\
        \hline
         \multicolumn{7}{c}{Through fibres}                                       \\
         \hline
    $\Phi_1$ & 0.000729  & 0.01277  & 7.464  & 0.1321  & 0.10191  & 0.09018 \\
    $\Phi_2$ & 0.002332  &          & 4.224  & 0.0371  & 0.10084  & 0.09088 \\
    $\Phi_3$ & 0.002685  & 0.00229  & 3.548  & 0.0628  & 0.08470  & 0.07404
    \end{tabular}
    \caption{Table of results from 1D attenuation modelling using BLR; $\mathbf{w}_1$ \& $\mathbf{w}_2$ are model weight parameters, $\sigma^2$ is the estimated variance of the function, $\sigma_{\mathbf{w}}$ is the estimated variance of the weights, $NMSE_{tr}$ is the \textit{negative mean squared error} between the model and the dataset used for training, and $NMSE_t$ is between the model and the independent validation set.}
    \label{tab:BLR_results}
\end{table}

As can be seen in \Cref{tab:BLR_results}, the confidence in the mean weights (which increases as $\sigma_{\mathbf{w}}$ decreases), is similarly large for model $\Phi_1$ but is much more confident for model $\Phi_2$ when being applied across the fibres. %
This observation may also form the basis for further investigation.

\subsection{Two-dimensional Gaussian process models}

In this section, the various approaches to modelling the two dimensional feature space summarised in \cref{tab:kernModels} will be compared. %
Models are compared visually and based upon a number of metrics as described in \cref{app:perfMetr}. %
The quantitative assessment of the models is discussed in \cref{sec:quant_mod}. %
These metrics are the log marginal likelihood $LML$, which is a measure of how well the model fits the training data. %
Next, the predictive log likelihood of an independent test set, considered in the case where every prediction is assumed independent $PLL_i$, and when the predictions are assumed correlated $PLL_c$. %
Finally, the normalised mean squared error $NMSE$ of the mean fit to the independent test set is also computed. %
This final metric should be treated with care, since it does not represent the quality of the uncertainty quantification in the model fit. %
The most rigorous test of these models can be considered to be the correlated predictive log likelihood which captures the full correlation of the predictive model including the mean, variance and covariance predictions. %
For all graphical representations shown, the data is presented in the $\log_{10}$ scale, but the models were all trained directly on the values of $h_m$. Therefore, the units for the figures are in $\log_{10}(mm)$. %

\subsubsection{Uninformed Gaussian process models}

\begin{figure}[h!]
    \begin{subfigure}{0.48\textwidth}
        \includegraphics[width=\textwidth]{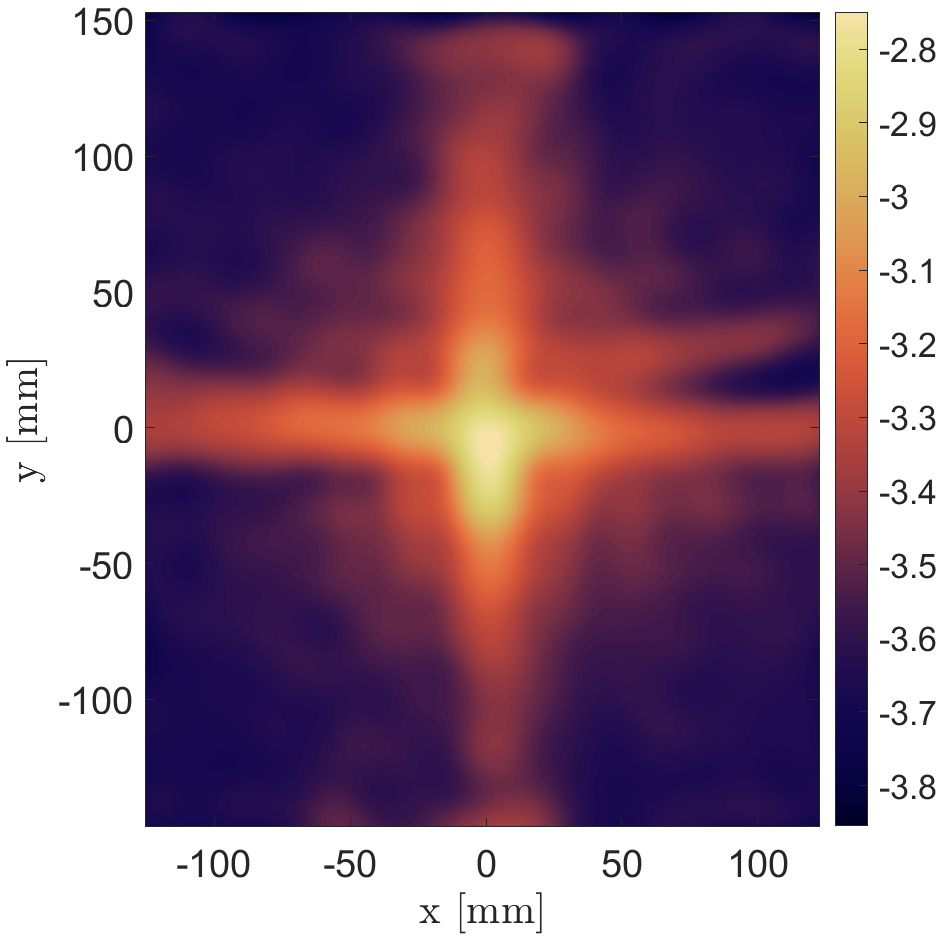}
        \caption{}
        \label{fig:GP_BB_Cart}
    \end{subfigure}
    \begin{subfigure}{0.48\textwidth}
        \includegraphics[width=\textwidth]{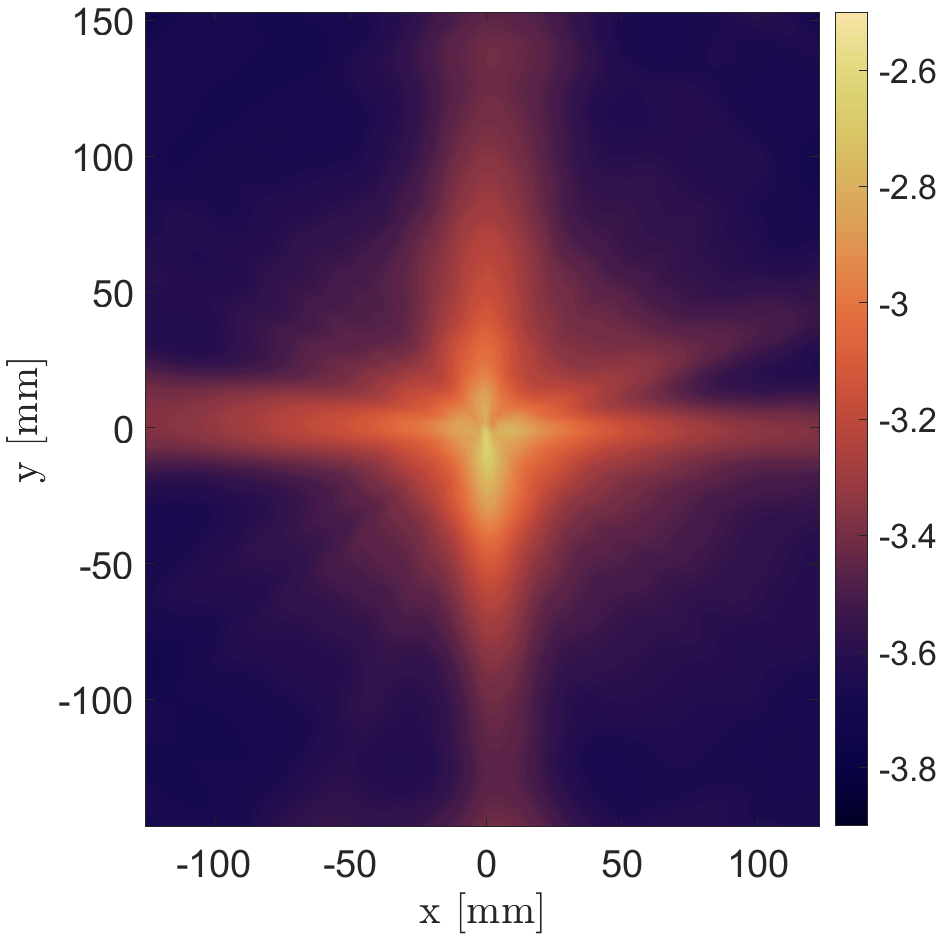}
        \caption{}
        \label{fig:GP_BB_Pol}
    \end{subfigure}
    \caption{Results of uninformed `black-box' GPs; (a) GP strategy A and (b) GP strategy B. The units are in $\log_{10}(mm)$.}
    \label{fig:GP_BB}
\end{figure}

\cref{fig:GP_BB} shows the mean predictions of the two uninformed GP models (A and B in \cref{tab:kernModels}). %
It can be clearly seen, that even without specific prior knowledge, the use of polar coordinate system (model B) offers a significant improvement over the Cartesian approach (model A). %
Model A appears to be an `out-of-focus' copy of the original data, whereas model B has generated a smoother function which is more likely to represent the physical mechanisms by which the wave operates. %
Even this simple consideration of the structure of the data being modelled leads to far more consistent results from the model. %
The quality of the fits for each of these models is compared qualitatively in \cref{tab:GP_results} along with the other GP models. %

Since the wave attenuation data naturally follows (approximately) a polar behaviour, one could envisage this problem as the GP trying to learn the mapping of Cartesian to polar spaces as well as the mapping from the polar to the feature-space. %
This two-stage mapping is being attempted through a single kernel and significantly complicates the modelling problem, thus, it is likely to underperform a model specified in the correct space. %
Further to this, it is concerning that the Cartesian model may have attempted to model some of the structure in the measurement noise. %
This is again a topic of further investigation for the future. %

However, as can be seen in \cref{fig:GP_BB_Pol}, by learning the model through a uninformed kernel operating on the polar coordinates, a more believable model of the feature space is learnt. %
These results show that even when implementing machine learning methods with no direct embedding of the physical process, the space in which the function operates must still be taken into account.

\subsubsection{Guided wave mean functions}

Models C and D in \cref{tab:kernModels} show the two approaches where mean function behaviour is included in the model to capture the expected behaviour of the guided waves. %
Both of these models make use of the third one-dimensional attenuation model from \cref{sec:BLR}, which includes both geometric spreading and viscoelastic damping.
Model C couples this mean behaviour in the radial direction with the flexible polar kernel used in model B, this allows the GP to learn functional behaviour in both the radial and angular dimensions. %
Model D is more restrictive and it is assumed that the mean function captures all of the radial behaviour and the GP only models functional behaviour in the angular dimension.

\begin{figure}[h]
    \begin{subfigure}{0.48\textwidth}
        \includegraphics[width=\textwidth]{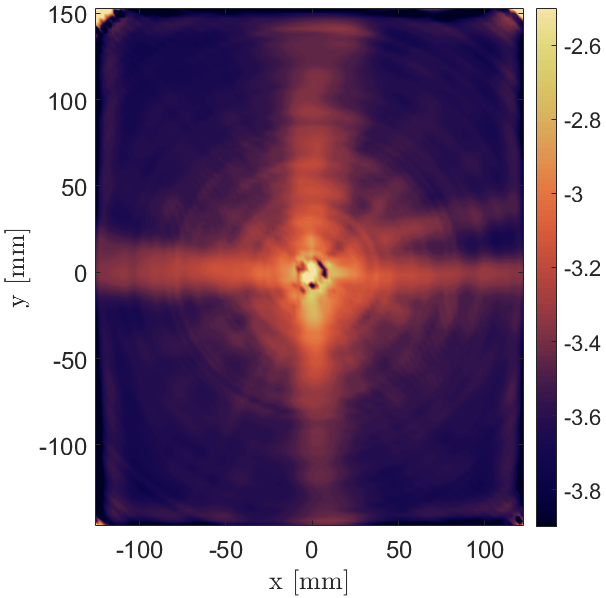}
        \caption{}
        \label{fig:GP_GB_mean_Cov}
    \end{subfigure}
    \begin{subfigure}{0.48\textwidth}
        \includegraphics[width=\textwidth]{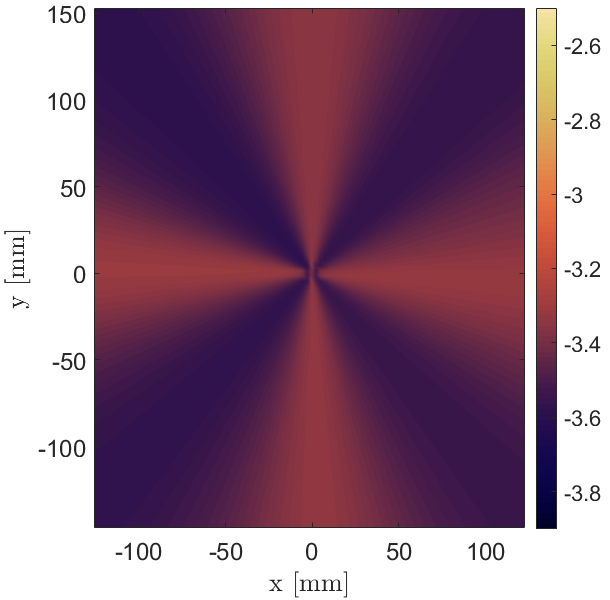}
        \caption{}
        \label{fig:GP_GB_mean_noCov}
    \end{subfigure}
    \caption{Results of informed model generated using generic kernels and inclusion of a mean function in the GP, with (a) model C, the full polar kernel, and (b) model D, the GP only modelling angular behaviour. The units are in $\log_{10}(mm)$.}
    \label{fig:GP_GB_mean}
\end{figure}

The resulting mean predictions on an independent test set for each of these models are shown in \cref{fig:GP_GB_mean}. %
Considering the prediction shown in \cref{fig:GP_GB_mean_Cov}, a `banding' effect is seen as a circular structure centred on the origin. %
This `banding' is most prominent in areas of propagation through the fibres and less prominent in the central region, around the wave source. %
This banding can be explained by considering how the mean function is included in the model. %
The mean function $m(\mathbf{x})$ is likely to fit the mean of the $\rho$ dimension well as the two dominant attenuation mechanisms captured. %
The inclusion of the mean function in a Gaussian process can be imagined as subtracting this function from the relevant dataset. %
When the mean function captures most of the behaviour, only unstructured data should be left to be modelled by the GP covariance, i.e.\ the noise in the system. %
In the results shown here, the mean function fits the data well and the unstructured data along the radial $\rho$ dimension is still modelled in \cref{eq:matern_pol}.
The GP can still attempt to find a structure in unstructured data. %
An interesting note from this result is that if there were functional information in the data still to be inferred, this would be picked up by the covariance kernel. %
In this case it is believed that the banding artefact may be due to the GP modelling structure in the noise on this realisation of the measurement, it is expected that if further training data were included, this effect would diminish.

To avoid this issue, and assuming that the mean function models well the radial attenuation behaviour, model D does not include the radial component in its covariance kernel, so it has covariance given by \cref{eq:wend_def}. %
The results of training the model with exclusion of the radial kernel can be seen in \cref{fig:GP_GB_mean_noCov}, in which it can be seen that the banding artefacts are no longer evident. %
However, the model also appears to lose accuracy as the value of $h_m$ attenuates much more quickly away from the source than is seen in \cref{fig:ogScan}. %
The loss of the banding artefacts demonstrates that it is the inclusion of this radial dimension in the kernel which leads to this phenomenon. %
These models will also be compared quantitatively once all models have been shown qualitatively. 

\subsubsection{Kernels capturing guided wave behaviour}

The final two models (E and F) attempt to embed understanding of the guided-waves by directly modifying the prior belief in the model through kernel design. %

\begin{figure}[h!]
    \centering
    \begin{subfigure}{0.48\textwidth}
        \includegraphics[width=\textwidth]{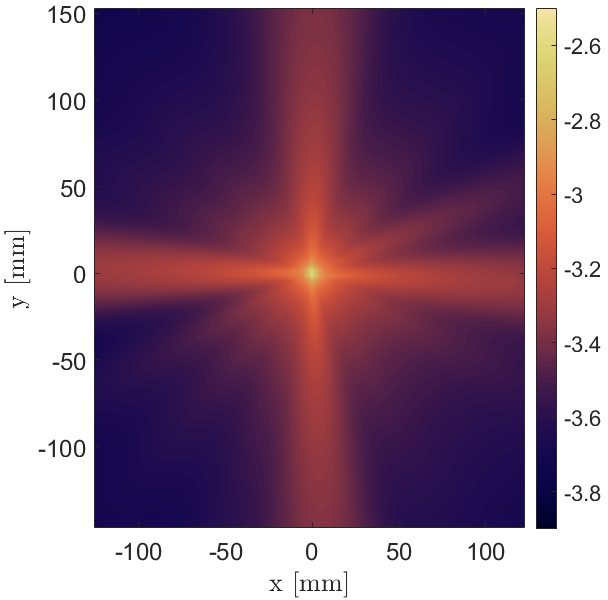}
        \caption{}
        \label{fig:kern1_GP_sp}
    \end{subfigure}
    \begin{subfigure}{0.48\textwidth}
        \includegraphics[width=\textwidth]{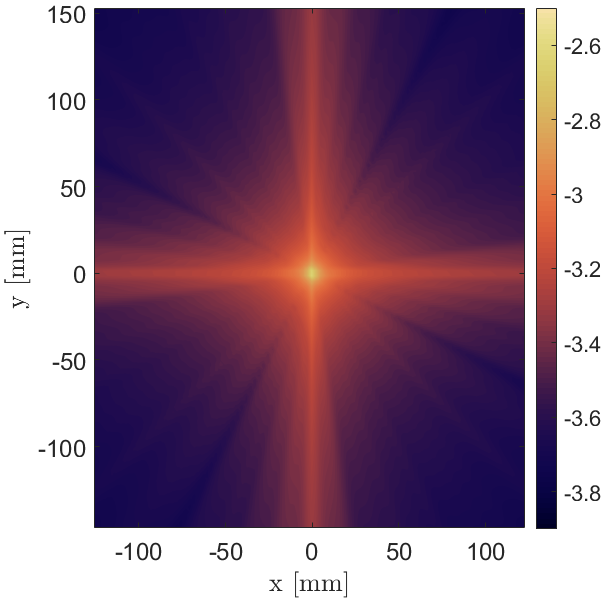}
        \caption{}
        \label{fig:kern2_GP_sp}
    \end{subfigure}
    \caption{Results of informed model generation using non-generic kernels, here indicating the expected mean function over the entire space when modelling with (a) GP strategy E and (b) GP strategy F.  The units are in $\log_{10}(mm)$.}
    \label{fig:GP_GB_kern}
\end{figure}

The estimated mean predictions on an independent test set for each of these models are shown in \cref{fig:kern1_GP_sp,fig:kern2_GP_sp}. %
It is clear in these results, how even the small changes between the two kernels can significantly impact the function space that is learnt. %
Comparatively, model E (\cref{fig:kern1_GP_sp}) leads to a much `smoother' result in comparison to model F (\cref{fig:kern2_GP_sp}). %
Fewer high-frequency components are seen in the angular dimension leading to this appearance.
This difference is due to the differences in the prior belief imposed in the angular kernels for each of these models. %

An interesting result seen here is the spreading of energy away from the fibre orientation. %
This spreading could be physically explained by acknowledging the secondary-guiding characteristics of the fibres themselves. %
As the waves travel along the fibres independently, they will lose some energy into adjacent media (i.e. the epoxy matrix). %
This will manifest itself as energy spreading in a different direction to the fibre orientation. %

Visually, kernel E appears to allow the capture of the spreading of the wave energy better.
In \cref{fig:kern2_GP_sp} it appears that the decay of energy is not captured well and that energy is only propagating along the fibres, not across. %
The periodicity enforced in model F is obvious to see in the significantly lower value `band' at approximately $\theta=\{20^{\circ},110^{\circ},200^{\circ},290^{\circ}\}$. 
The kernel used in model E offers greater flexibility in symmetry as a result of the additive combination of the SE kernel along the angular dimension. %
For both kernels, the predictive mean has less variation in the function modelling wave propagation along the fibres, and is a less smooth function when propagating through the fibres. %
This difference could be improved by increasing the signal-to-noise ratio of the experiment.
This alteration will help model areas of high attenuation (i.e.\ low energy); since the energy of the wave decreases significantly away from the fibre orientation, the value of $h_m$ may not exceed the noise floor. %()
Currently, in regions with large $\rho$ the data becomes unstructured and it is difficult to infer the function with as much confidence. %
It may also be worth exploring a heteroscedastic noise model in future work \cite{Kersting2007}.

\begin{figure}[h!]
    \centering
    \begin{subfigure}{0.48\textwidth}
        \includegraphics[width=\textwidth]{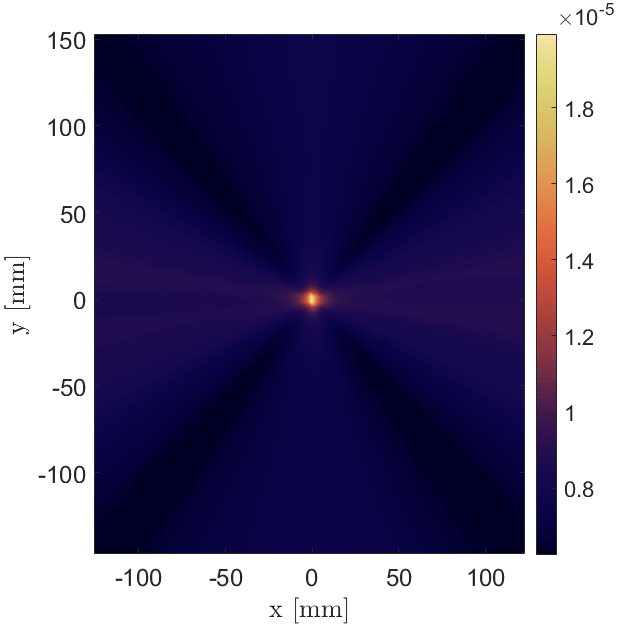}
        \caption{}
        \label{fig:kern1_GP_var}
    \end{subfigure}
    \begin{subfigure}{0.48\textwidth}
        \includegraphics[width=\textwidth]{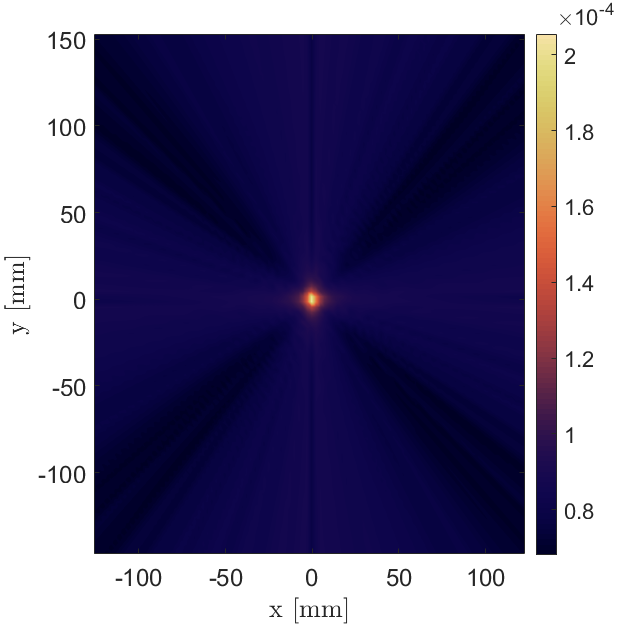}
        \caption{}
        \label{fig:kern2_GP_var}
    \end{subfigure}
    \caption{Results of informed model generation using non-generic kernels, here indicating the expected variance over the entire space when modelling with (a) GP strategy E and (b) GP strategy F.}
    \label{fig:GP_GB_kern_var}
\end{figure}

For both models E and F, the estimated variance over the surface generates similar results; this can be seen in \cref{fig:kern1_GP_var,fig:kern2_GP_var}. %
Both models see a sharp increase in variance towards the centre, this can be explained by examining the one-dimensional attenuation kernel proposed in \cref{eq:rad_kernel}. %
The polynomial kernel \cref{eq:rad_kernel} included in these models will result in functions that tend to infinity with $\rho \rightarrow0$. %
This kernel is used in multiplicative combination with the exponential decay kernel \cref{eq:expDec_kernel}, resulting functions with the same characteristic. %
The second kernel design results in a slight increase in variance at approximately $\theta=\{50^{\circ},140^{\circ},230^{\circ},320^{\circ}\}$. %
From physical interpretation of how fibres affect the energy of the waves, the variance should not increase at this orientation if not also at $\theta=\{40^{\circ},130^{\circ},\textrm{etc.}\}$. %
For both kernel designs, there also appears a greater variance in the energy of the wave when propagating along the fibres; this is likely a result of the short range of $\theta$ in which the wave has directly propagated away from the source along a single fibre. 

\subsection{Quantification of model performance}
\label{sec:quant_mod}

Until now, the modelling approaches for guided wave features have been compared in a qualitative manner. %
It has been discussed how certain models give rise to desirable characteristics in the latent functions being learnt, which may obscure the physical behaviour. %

\begin{table}[h!]
    \centering
    \small
    \begin{tabular}{ c | c | c | c | c | c | c }
         Model & $m(\mathbf{x})$ & $k(\mathbf{x},\mathbf{x}')$  & $LML$ & $PLL_i$ & $PLL_c$ & $NMSE$ \\
         \hline
         \hline
         A & 0 & $k_{\textrm{SQE}}(\{x,y\},\{x,y\}')$  & 24,301 & 3,745.1 & 3,770.9
 & 4.424 \\
         B & 0 & $k_{2}(\{\rho,\theta\},\{\rho,\theta\}')$ & 50,012 & 9,964.4 & 9,975.40 & \textbf{4.0567} \\
         C & $A_3(\rho)$ & $k_{2}(\{\rho,\theta\},\{\rho,\theta\}')$ & 19,872 & 13,047.7 & 4,221.4 & 1,037.7 \\
         D & $A_3(\rho)$ & $k_{\textrm{W}}(\theta,\theta')$ & 14,950 & 11,176.6
 & 3,326.9 & 29.9 \\
         E & 0 & $k_3(\{\rho,\theta\},\{\rho,\theta\}')$ & \textbf{96,952} & \textbf{21,013.6} & \textbf{22,727.4} & 9.9422 \\
         F & 0 & $k_4(\{\rho,\theta\},\{\rho,\theta\}')$ & 75,682 & 16,274.9 & 16,291.3 & 9.9133 \\
    \end{tabular}
    \caption{Table of qualitative assessment values for 2D GP modelling strategies tested, indicating the \textit{log-marginal likelihood} (LML), \textit{independent predictive log likelihood} ($PLL_i$), \textit{co-dependent predictive log likelihood} ($PLL_c$), and \textit{normalised mean square error} (NMSE). Best values for each metric are highlighted in bold.}
    \label{tab:GP_results}
\end{table}

\cref{tab:GP_results} shows the results of the GP models tested against the metrics previously described. %
These metrics allow insight into the accuracy and validity of resulting models, and also provides an opportunity to discuss what might be meant by a ``good model''. %
It is important to consider that any given model is only as good as what it will be ultimately used for. %
In the context of engineering, specifically in  NDE and SHM, these models will be used to make operational decisions about the system. %
As such, it should be considered whether the end user is most interested in the point-wise prediction accuracy in which case the NMSE is the most appropriate metric. %
If instead, the models will be carried forward into a risk-driven assessment, then capturing the full uncertainty in the model is important and the correlated predictive log likelihood will be the most appropriate metric. %
In this work, a number of possible assessment criteria are presented; the onus is on the end user to choose the model which best captures the characteristics of the data/function which are most important to them. %
It is the opinion of the authors that the most robust measure of how well the functional behaviour of the feature space is captured is the correlated predictive log likelihood $PLL_c$. %

Turning attention to specific results from these experiments, the results for each of the six models under each metric are shown in \cref{tab:GP_results}. %
It can be seen that models A and B result in the lowest NMSE scores, in other words that their point-wise predictions are closest to the observed test data. %
This result may be expected since they are the most flexible models. %
Interestingly, model B, the informed polar coordinate model, also recovers the highest independent predictive log likelihood. %
Interpreting this result; if one wants to predict only the behaviour at single points on the plate and is not concerned with the correlation between these predictions, then this is the optimal model (from those tested). %
It is also seen in the log marginal likelihood $LML$ that model B captures the behaviour in the training data much better than model A. %

The inclusion of the guided-wave attenuation models as mean functions in the GP (models C and D) appears not to produce desirable effects. %
These models perform worst in their representation of the training data $LML$ and their point-wise prediction capability $NMSE$. %
For this dataset, this is compelling evidence that the inclusion of the guided wave attenuation mechanisms through a mean function does not lead to a useful model. %
As discussed, the models may be finding too much structure in the noise of the data used for training, especially model C. %
This postulation is evident in the very high $NMSE$ score, which would indicate that the model performs considerably worse than taking the mean of the prediction data. %
However, it is also important to note the exceptionally large NMSE for model C (representing an error of $\sim$1038\%), which is a result of the estimated mean function resulting in a 'singularity' towards the centre of the plate; $m(\mathbf{x}) \rightarrow \infty$ as $\rho \rightarrow 0$.

Finally, considering models E and F, where the knowledge of the guided waves is used to modify the prior belief in the model, via the covariance kernel; both of these models perform very well when considering their ability to model the training data assessed through the $LML$. %
When examining their performance on the independent test set, it is seen that the $NMSE$ score is around 9\% compared to the 4\% of models A and B; this along with their poor independent predictive log likelihoods $PLL_i$ are indicators that the point-wise predictions from these models are not as good as models A and B. %
However, in terms of capturing the complete function space they far exceed all the other modelling strategies. %
These approaches appear to best capture the underlying functional behaviour of the guided-wave attenuation. %
For this reason, these models can be considered to be the most suitable for de-noising or spatially up-sampling the data. %
These two approaches may also be considered the most robust methodologies for modelling the guided-wave behaviour not currently described by governing equations, e.g.\ outlying behaviour.
This advantage has been achieved through incorporation of the physical mechanisms driving the attenuation in the prior specification of the model, via modification of the covariance kernel.

The results from this dataset and model would support the use of model E for modelling the feature-space of $h_m$ for a guided wave in a CFRP plate. %
However, when using this strategy users should consider all models and the system being modelled, as well as the level of physical knowledge that is currently available for said system. 

\section{Conclusion and further work}
\label{sec:conclusion}

As computation capabilities and machine learning techniques are becoming increasingly more accessible, the adoption of such methods to solve engineering problems is also becoming more prominent. This increasing use of such methods can present many underlying issues with the results, such as unreasonable assumptions, lack of extrapolation capability and computational costs. However, by implementing physical knowledge to guide learning, more robust models may be generated which can reduce many of these issues.

A barrier to the progression of using guided waves in an NDE or SHM strategy is the difficult of modelling the behaviour of these waves in complex materials. The work presented here shows promising steps towards generating a physics-incorporated, data-driven model for the feature-space of guided-waves in such materials. Several characteristics of such a strategy, which must be carefully considered to maintain robustness, have been discussed. This strategy provides a key framework for the development of guided-wave models for complex materials --- such as the ones used in this paper --- by allowing modelling of features which define the waves propagating throughout the material. The important distinction of this combined method, in comparison to physics or data-driven-only methods, is that this allows physics to guide the model, whilst allowing unknown or undescribed physical mechanisms to be incorporated through the data-driven aspect. 

When initially looking at the kernels chosen to represent different learning strategies and levels of constraint, it is not clear which strategy will result in the optimal model output. By various qualitative assessment values it is possible to see how each model fits in comparison to the validation data, in different aspects. By leaving the model uninformed, it is possible to get a closer fit to the training and validation data in terms of difference between the predicted and measured values. However, by guiding the learning process using physics-based implementation of the problem, it is possible to get a higher likelihood model.

The work here has been shown for the case of modelling energy-based features of a guided wave in a CFRP plate structure, and though these specific kernels cannot be directly applied to some other complex scenarios --- such as complex geometry, quasi-periodic materials, or other features such as \emph{time-of-flight} --- the kernels can be modified in the framework presented here to model such systems. The structure is applied through the kernels to embed prior belief of the shape of the features over the topology being modelled.

Further work will be done to embed structural knowledge into the modelling process --- such as plate thickness, joints and layup information --- in order to improve extrapolation of data over guides of varying shape.

\section{Acknowledgements}

The authors gratefully acknowledge the support of the UK Engineering and Physical Sciences Research Council (EPSRC) [grant numbers EP/R004900/1, EP/R003645/1, EP/S001565/1, EP/J013714/1 and EP/N010884/1].

\begin{appendices}
\crefalias{section}{appendix}

\section{Bayesian Linear Regression}
\label{app:BLR}
\setcounter{equation}{0}
\renewcommand{\theequation}{\thesection.\arabic{equation}}

Traditional linear regression formulates a model using point estimates of a set of parameters which ``best'' fit an available dataset, based on minimising an $L^2$-norm between the model predictions and the data. %
Instead, BLR aims to establish a probability distribution of possible model parameters. %
The model has the form,
\begin{equation}
    y = \mathbf{w}^{\top} \phi(\mathbf{x}) + \varepsilon, \qquad \varepsilon \sim \mathcal{N} (0,\sigma^2)
    \label{eq:BLR_modelForm}
\end{equation}
where $\phi$ is some basis for expansion of a $p$-dimensional data point $\mathbf{x}$; $\mathbf{w} = \{ w_1,w_2, ... ,w_p \}$ are the associated weights of the basis expansion, and $\varepsilon$ is an additive Gaussian white noise distributed as $\mathcal{N}(0,\sigma^2)$. %
The weights $\mathbf{w}$ and the variance $\sigma^2$ are the unknown. %
The Bayesian linear regression model approach was chosen since it returns a quantified uncertainty. 
The task is then to compute the posterior distribution of the parameters $p(\mathbf{w},\sigma^2|D)$. %
This posterior distribution has the following form,
\begin{equation}
	p(\mathbf{w},\sigma^2|\mathcal{D}) = NIG(\mathbf{w},\sigma^2|\mathbf{w}_N,\mathbf{V}_N,a_N,b_N)
\end{equation}
with,
\begin{equation}
	\mathbf{w}_N = \mathbf{V}_N(\mathbf{V}_0^{-1}\mathbf{w}_0+\mathbf{X}^{\top}\mathbf{y})
\end{equation}
\begin{equation}
	\mathbf{V}_N = (\mathbf{V}_0^{-1}+\mathbf{X}^{\top}\mathbf{X})^{-1}
\end{equation}\begin{equation}
	a_N = a_0 +n/2
\end{equation}\begin{equation}
	b_N = b_0 +\frac{1}{2}\left(\mathbf{w}_0^{\top} \mathbf{V}_0^{-1} \mathbf{w}_0 + \mathbf{y}^{\top}\mathbf{y} - \mathbf{w}_N^{\top} \mathbf{V}_N^{-1}\mathbf{w}_N \right)
\end{equation}
where $\mathbf{V}_0$, $\mathbf{w}_0$, $a_0$ and $b_0$ are hyperparameters of the prior. %
It is possible to set an uninformative prior for $\sigma^2$ by applying $a_0=b_0=0$. %
Also setting $\mathbf{w}_0 = 0$ and $\mathbf{V}_0 = g(\mathbf{X}^{\top}\mathbf{X})^{-1}$ for any positive value $g$; leads to Zellner's \textit{g-prior} \cite{Zellner1986}. %
By having the prior variance proportional to $(\mathbf{X}^{\top}\mathbf{X})^{-1}$, it is ensured that the posterior is invariant to scaling of the inputs.

\section{Gaussian Process}
\label{app:GP}
\setcounter{equation}{0}
\renewcommand{\theequation}{\thesection.\arabic{equation}}

Conceptually, one can think of this process as estimating, rather than one ``best'' fit through the data, a distribution over all the possible functions that could explain the data. %
By virtue of its construction, the marginal and conditional distributions of any finite subset of data points in the function is Gaussian. %
In other words, any set of data modelled by the Gaussian process can be represented by a multivariate Gaussian distribution. %
The benefit of this result is that computations are normally available in closed form; for example, the conditional distribution of some new test points given the already observed data can be recovered exactly. %
The model is also \emph{nonparametric}; the form of the function which will fit the data does not need to be specified, i.e.\ it is not necessary to choose a basis, such as a polynomial one. %
Instead, the function is modelled by representing the covariance in the data through a kernel or covariance function. %
This kernel is used to embed belief about which \emph{family of functions} the data have come from, e.g.\ a nonlinear or periodic function.

The Gaussian process can be used to model nonlinear regression problems of the form,
\begin{equation}
    \mathbf{y} = f(\mathbf{X})+\boldsymbol{\varepsilon} \qquad \boldsymbol{\varepsilon} \sim \mathcal{N}(\mathbf{0},\sigma_n^2\mathbb{I})
\end{equation}
where $\mathbf{y}$ is a vector of $N$ observed targets, $\mathbf{X}$ a matrix of $N$ observed inputs in $D$ dimensions, and $\varepsilon$ a vector of realisations from a zero-mean Gaussian white noise process with variance $\sigma_n^2$.

A GP is fully defined by its mean and covariance function,
\begin{equation}
    f(\mathbf{x})\sim \mathcal{GP}\left(m(\mathbf{x}),k(\mathbf{x},\mathbf{x}')\right)
    \label{eq:GPdef}
\end{equation}

The mean function, $m(\mathbf{x})$ can be any parametric mapping of $\mathbf{x}$, e.g.\ a polynomial. %
The correlation between the targets is captured by the covariance function which expresses the similarity between two input vectors $\mathbf{x}$ and $\mathbf{x}'$. %
To predict at a new test point $\mathbf{x}_\star$, or set of test points $X_\star$, predictive equations are used to determine the expected mean function $\mathbb{E}[f_\star]$ and expected covariance $\mathbb{V}[f_\star]$ \cite{Rasmussen2005},
\begin{subequations}
    \begin{align}
    f_\star&\sim\mathcal{N}\left(\mathbb{E}[f_\star],\mathbb{V}[f_\star]\right) \label{eq:GP_predFunc}\\
    \mathbb{E}[f_\star] &= m(\mathbf{x}_\star)+k(\mathbf{x}_\star,X)(k(X,X)+\sigma_n^2\mathbb{I})^{-1}\mathbf{y} \label{eq:GP_predMean}\\
    \mathbb{V}[f_\star] &= k(\mathbf{x}_\star,\mathbf{x}_\star)-
    k(\mathbf{x}_\star,X)(k(X,X)+\sigma_n^2\mathbb{I})^{-1}
    k(X,\mathbf{x}_\star) \label{eq:GP_predVar}
    \end{align}
\end{subequations}

If predicting at noisy output locations, i.e. $y_\star$, it is trivial to add the noise variance $\sigma_n^2\mathbb{I}$ to the predictive covariance in \cref{eq:GP_predVar}. %
As such the GP returns the posterior distribution over $f_\star$ or $y_\star$ as a Gaussian distribution. %

For practical implementation of a Gaussian process, the reader is recommended to follow the guidance of Rasmussen \cite{Rasmussen2005}. %
The primary influence of the user when implementing a GP is in the choice of the kernel, which is calculated as any other kernel; linear pair-wise distances between points to form a covariance matrix. %
Careful consideration of data should also be applied in implementation, such as data type (scale, sign, etc.), data size and space on which it operates. %

Standard practice to determine the hyperparameters of a Gaussian process is to maximise the \emph{marginal likelihood}, which in practice is done by minimising the \emph{log marginal likelihood} (NLML),

\begin{equation}
    \hat{\mathbf{\Theta}}= \textrm{arg min}(-\log p(\mathbf{y}\vert\mathbf{\Theta}))
\end{equation}

\noindent where the negative log marginal likelihood of the Gaussian process is given by,

\begin{equation}
    -\log p(\mathbf{y}\vert\mathbf{X},\mathbf{\Theta}) = \frac{1}{2}\log\vert K_y\vert +\frac{1}{2}\mathbf{y}^{\top}K_y^{-1}\mathbf{y} + \frac{N}{2}\log(\sigma_n^2)
    \label{eq:nlml}
\end{equation}

When defining $K_y = K(X,X)+\sigma_n^2\mathbb{I}$, $K(X,X)$ is the pairwise covariance matrix of all of the training inputs and $N$ is the number of data points. %

\section{Polar GP}
\label{app:polarGP}
\setcounter{equation}{0}
\renewcommand{\theequation}{\thesection.\arabic{equation}}

To make this modification is not as trivial as it may seem. %
Remembering that the covariance function is a measure of similarity between two data points it is necessary to define a kernel which encodes this. %
Specifically, it is necessary to have high covariance between points that are close to each other in angle. %
For example a point with angle $359^\circ$ should have a high covariance with $1^\circ$ if the radii are also close. %
This will require modifications to the kernel in terms of the distance used to assess how close points are together and also to the covariance function itself. %
Padonou and Roustant \cite{Padonou2015} suggest two potential definitions for a distance which fulfils this criteria, full details of setting up a polar coordinates kernel can be found in that work but it is briefly reviewed here. %
These two distances are: the \emph{chordal} distance $ d_1(\theta,\theta ') = 2\sin\left(\frac{\theta-\theta '}{2}\right) $ or the \emph{geodesic} distance $ d_2(\theta,\theta ') = \arccos(\cos(\theta-\theta ')) $. %
Using these definitions, it is possible to define the covariance in the $\theta$ dimension of a $\left\{\rho,\theta\right\}$ polar coordinate system. %
The $C^2$-Wendland function is used as the kernel, since this produces a covariance of $0$ when $d_2=\pi$ and is strictly positive when $d_2>\pi$, both necessary conditions for the polar kernel design. %
The $C^2$-Wendland function is defined as,
\begin{equation}
	W_c(t) = \left(1+\tau\frac{t}{c}\right) \left(1-\frac{t}{c}\right)^{\tau}_+ , \quad c \in [0,\pi]; \tau \ge 4
	\label{eq:wend_def}
\end{equation}
When applying the Wendland function as the covariance function, the value of $c$ must change depending on the angular distance chosen,
\begin{equation}
k_{\textrm{W}} = \begin{cases}
	k_{chord}(\theta,\theta ') = W_2(d_1(\theta,\theta ')) \\
	k_{geo}(\theta,\theta ') = W_{\pi}(d_2(\theta,\theta '))
	\end{cases}
	\label{eq:cov_ang}
\end{equation}
Here, the value of $\tau$ acts as a `steepening' parameter on the angular covariance; this can be seen as the angular analogue to the inverse of the length scale parameter described previously. %

To form a full polar covariance function, a different kernel is applied only on the radial dimension of the input. %
This kernel could be any stationary isotropic kernel; in this work the Mat\'ern 5/2 kernel is used as in \cite{Padonou2015}. %
In that case the distance used in the Mat\'ern kernel is the absolute difference between the two radial components $\vert\rho-\rho^\prime\vert$. %
For the angular component, \cref{eq:wend_def} is used with the geodesic distance such that kernel $k_{geo}$ is used.

These choices define the covariance in the model along each of the directions - the radial $\rho$ and the angular $\theta$. %
To form the total covariance it is necessary to combine these two. %
It is known that the addition or pointwise multiplication of any two valid covariance functions is itself a valid covariance \cite{Rasmussen2005}. 
In this work an \textit{ANOVA} combination \cite{Wahba1990} of the kernels in each dimension is used, as in \cite{Padonou2015}. %
This allows variations in each dimension as well as the combination to contribute to variation in the function. %
The combined \textit{ANOVA} kernel is defined as,
\begin{equation}
	k_2(\mathbf{x},\mathbf{x}') = \sigma_f^2\left(1+\sigma_{f,r}^2k_{\textrm{mat}}(\rho,\rho')\right)\left(1+\sigma_{f,a}^2k_{\textrm{W}}(\theta,\theta ')\right)
	\label{eq:anova_kern}
\end{equation}
\noindent where $\sigma_{f,m}$ and $\sigma_{f,a}$ act as weights representing the influence of changes in each dimension on a change in the output.

\section{Kernels}
\label{app:kernels}
\setcounter{equation}{0}
\renewcommand{\theequation}{\thesection.\arabic{equation}}

Many of the kernels shown in this paper use the distance between data points, $r$, for their calculation; for this paper, $\mathbf{x} - \mathbf{x} '$ is the Euclidean distance when data is in Cartesian coordinates or applied to $\rho$ in polar coordinates and is defined by $d_2=\arccos(\cos(\theta-\theta'))$ when applied to $\theta$ in polar coordinates.

\subsection{Squared Exponential}

The squared exponential is a general nonlinear kernel which operates in a $D$-dimensional real space, where $r$ is the distance between two points $\mathbf{x}$ \& $\mathbf{x}'$. 

\begin{equation}
    k_{\textrm{sqe}}(\mathbf{x},\mathbf{x}') = \sigma_f^2 \exp\left(-\frac{r^2}{2l^2}\right)
\end{equation}

\paragraph{Hyperparameters} $\ell$ is the length scale, $\sigma_f^2$ is the scaling factor.

\subsection{Mat\'ern Class}

The Mat\'ern class of kernels are general nonlinear kernels which operate in a $D$-dimensional real space, and are specified by a scaling factor $\nu$; in this case, $\nu =5/2$ is used,

\begin{equation}
    k_{\textrm{mat},\nu=5/2}(\mathbf{x},\mathbf{x}') = \sigma_f^2\left(1+\frac{\sqrt{5}r}{l}+\frac{5r^2}{3l^2}\right) \exp\left(-\frac{\sqrt{5}r}{l}\right)
\end{equation}

\paragraph{Hyperparameters} $\ell$ is the length scale, $\sigma_f^2$ is the scaling factor.

\subsection{Polynomial}

The polynomial kernel is an inhomogeneous linear kernel which operates in a $D$-dimensional real space, and is specified by the coefficient $p$,

\begin{equation}
    k_{\textrm{pol}}(\mathbf{x},\mathbf{x}') = \sigma_f^2(\mathbf{x}\cdot\mathbf{x}'+\sigma_0^2)^p
\end{equation}

\paragraph{Hyperparameters} $\sigma_0^2 \geq 0$ is the offset term trading off the influence of higher-order versus lower-order terms, $\sigma_f^2$ is the scaling factor.

\subsection{$C^2$-Wendland}

The $C^2$-Wendland kernel is a stationary kernel which operates on the $\theta$-dimension of a real polar space, and was shown for use in Gaussian process modelling in \cite{Padonou2015}. $c=2$ when $d=d_1(\theta,\theta')=2\sin\left(({\theta-\theta'})/{2}\right)$ and $c=\pi$ when $d=d_2(\theta,\theta')=\arccos(\cos(\theta-\theta'))$,

\begin{equation}
    k_{\textrm{W}}(\theta,\theta') = \sigma_f^2\left(1+\tau\frac{d}{c}\right)\left(1-\frac{d}{c}\right)^{\tau}
\end{equation}

\paragraph{Hyperparameters} $\tau \geq 4$ is the variance steepening parameter, $\sigma_f^2$ is the scaling factor.

\subsection{Exponential Decay}

The exponential decay kernel is a non-general linear kernel which operates in a $D$-dimensional real space,

\begin{equation}
    k_{\textrm{ed}}(\mathbf{x},\mathbf{x}') = \sigma_f^2\exp\left(-\mathbf{x}l\right)\cdot\exp\left(-\mathbf{x}'^{\top}l\right)
\end{equation}

\paragraph{Hyperparameters} $\ell$ is the length scale, $\sigma_f^2$ is the scaling factor.

\subsection{Strictly Periodic}

The strictly-periodic kernel is a non-general linear kernel which operates on the $\theta$-dimension of a real polar space; the nature is enforced by the number of equally distributed symmetry axes, $n$,

\begin{equation}
    k_{\textrm{sym}}(\theta,\theta') = \sigma_f^2\left(\alpha_1+\alpha_2 \cos\left(2nd_2(\theta,\theta')\right)\right)
\end{equation}

\paragraph{Hyperparameters} $\alpha_1$ \& $\alpha_2$ are the relative weighting of the maximum value and periodicity of the function respectively.

\subsection{Generic Polar}

The generic polar kernel is an ANOVA combination of $k_{\textrm{mat}}$ and $k_W$ which operates on the $\{\theta,\rho\}$-dimension real polar space,

\begin{equation}
    k_2(\mathbf{x},\mathbf{x}') = \sigma_f^2\left(1+\sigma_{f,r}^2k_{\textrm{mat}}(\rho,\rho')\right)\left(1+\sigma_{f,a}^2k_{\textrm{W}}(\theta,\theta ')\right)
\end{equation}

\paragraph{Hyperparameters} $\sigma_{f,r}^2$ \& $\sigma_{f,a}^2$ are the relative importance of changes in the radial \& angular dimension respectively, $\sigma_{f}^2$ is the overall scaling factor.

\subsection{Angular Informed}

The angular informed kernel is an OR combination of $k_{\textrm{sqe}}$ and $k_{\textrm{sym}}$ which operates on the $\theta$-dimension of a real polar space,

\begin{equation}
    k_{\textrm{ang}}(\mathbf{x},\mathbf{x}') = \sigma_{f,\textrm{sqe}}^2 k_{\textrm{sqe}}(\theta,\theta') + \sigma_{f,\textrm{sym}}^2k_{\textrm{sym}}(\theta,\theta')
\end{equation}

\paragraph{Hyperparameters} $\sigma_{f,\textrm{sqe}}^2$ \& $\sigma_{f,\textrm{sym}}^2$ are the relative scaling factors for each kernel.

\subsection{Radial Informed}

The radial informed kernel is an AND combination of $k_{\textrm{pol}}, \; p=-1/2$ and $k_{\textrm{ed}}$ which operates on the $\rho$-dimension of a real polar space,

\begin{equation}
    k_{\textrm{rad}}(\mathbf{x},\mathbf{x}') = \sigma_{f,r}^2 k_{\textrm{pol}}(\rho,\rho') \; \cdot \; k_{\textrm{ed}}(\rho,\rho')
\end{equation}

\paragraph{Hyperparameters} $\sigma_{f}^2$ is the overall scaling factor.

\subsection{Informed Guided Wave, smooth \& strictly periodic}

The first UGW-informed kernel is an ANOVA combination of $k_{\textrm{ang}}$ and $k_{\textrm{rad}}$ which operates on the $\{\theta,\rho\}$-dimension of a real polar space. The periodicity in $\theta$ is enforced and smooth,

\begin{equation}
    k_3(\mathbf{x},\mathbf{x}') = \sigma_{f}^2\left(1+\sigma_{f,a}^2k_{\textrm{ang}}(\mathbf{\theta},\mathbf{\theta}')\right)\left(1 + \sigma_{f,r}^2k_{\textrm{rad}}(\mathbf{\rho},\mathbf{\rho}')\right)
\end{equation}

\paragraph{Hyperparameters} $\sigma_{f,a}^2$ \& $\sigma_{f,r}^2$ are the relative importance of changes in the angular \& radial dimension respectively, $\sigma_{f}^2$ is the overall scaling factor.

\subsection{Informed Guided Wave, smooth \& strictly periodic}

The second UGW-informed kernel is an ANOVA combination of $k_{\textrm{W}}$ and $k_{\textrm{rad}}$ which operates on the $\{\theta,\rho\}$-dimension of a real polar space. The periodicity in $\theta$ is enforced but not necessarily smooth,

\begin{equation}
    k_4(\mathbf{x},\mathbf{x}') = \sigma_{f}^2\left(1+\sigma_{f,a}^2k_{\textrm{W}}(\mathbf{\theta},\mathbf{\theta}')\right)\left(1 + \sigma_{f,r}^2k_{\textrm{rad}}(\mathbf{\rho},\mathbf{\rho}')\right)
\end{equation}

\paragraph{Hyperparameters} $\sigma_{f,a}^2$ \& $\sigma_{f,r}^2$ are the relative importance of changes in the angular \& radial dimension respectively, $\sigma_{f}^2$ is the overall scaling factor.

\section{Performance Metrics}
\label{app:perfMetr}
\setcounter{equation}{0}
\renewcommand{\theequation}{\thesection.\arabic{equation}}

It is necessary at this point to develop some metrics by which the models can be assessed. %
For this purpose, the experimental dataset was split into a training set, $\mathbf{x}_t$ and a test set $\mathbf{x}_{v}$, which consisted of 75\% and 25\% of the total dataset respectively. %
The performance of each model is reported on both the training and the test data; it is important to consider the test data performance, as this is the best indicator of which models are able to \emph{generalise}, i.e.\ which will work best on new unseen data.

The first metric used here is the \textit{normalised mean squared error} (NMSE) which can be computed for both training set ($NMSE_{tr}$) and test set ($NMSE_t$). %
For descriptive purposes, the NMSE indicates how well the estimate of the output fits with the observed values. %
The NMSE will return a score of zero when the predicted values are identical to those measured (this is impossible in the presence of any noise). %
A score of 100 is equivalent to simply taking the mean of the observed data as the prediction at every instance. %
The calculation for the NMSE is,

\begin{equation}
    NMSE = \frac{100}{n \sigma_y^2} {\left(\mathbf{y} - \mathbf{y}^\star \right)^{\top}\left(\mathbf{y} - \mathbf{y}^\star \right)} 
    \label{eq:NMSEt}
\end{equation}

\noindent where $\mathbf{y}$ is the vector of observed outputs and $\mathbf{y}^\star$ the predicted outputs (in this case the mean of the predictive distribution from a Gaussian process). %
$n$ is the number of observations in $\mathbf{y}$ and $\sigma_y^2$ the variance of those observations.

The second metric will be to compare the predictive likelihoods of the model. %
This metric can be a more informative way of assessing the models as it takes into account the uncertainty in the prediction as well as the quality of the mean fit. %
The predictive likelihood is given as $p\left(\mathbf{y}^\star \vert \mathbf{x}^\star,\mathbf{y},\mathbf{x}\right)$; this will change dependent upon the model being assessed but since all models in this work have a tractable Gaussian posterior it is given here by,
\begin{equation}
    p\left(\mathbf{y}^\star \vert \mathbf{x}^\star,\mathbf{y},\mathbf{x}\right) = \mathcal{N}\left(\mathbb{E}\left[\mathbf{y}^\star\right],\mathbb{V}\left[\mathbf{y}^\star\right]\right)
\end{equation}
This work will use this quantity in two ways; the first will consider each prediction to be independent, by not including the cross covariance terms in $\mathbb{V}\left[\mathbf{y}^\star\right]$ the predictive variance matrix. %
The full covariance of the prediction will also be considered from the Gaussian process, as this can give better insight into how well the function has been modelled.

To ensure computational stability these likelihood estimates are both calculated in the log space. %
The first quantity will be referred to as the independent predictive log likelihood $PLL_i$, and is defined by,
\begin{equation}
    PLL_i = \sum_{i}^{N}\log \mathcal{N}(\mathbf{y}_{i}\vert\mathbb{E}[\mathbf{y}_{i}],\mathbb{V}[\mathbf{y}_{i}],\mathbf{\Theta})
\end{equation} 
\noindent for $N$ data points. This is the product over the predictive likelihoods for every point, i.e.\ the joint likelihood if they were uncorrelated. %
The second will be considered the co-dependent predictive log likelihood $PLL_c$, defined by, 
\begin{equation}
    PLL_c = \log p(\mathbf{y}|\mathbb{E}[\mathbf{y}],\mathbb{V}[\mathbf{y}],\mathbf{\Theta})    
\end{equation}
where $PLL_c$ is computed as the likelihood of the full multivariate Gaussian over the predictive points, including the predicted covariance between those points.

\end{appendices}

\bibliography{UGW_GP_JSV_Ref}

\end{document}